\documentclass[10pt]{article} % For LaTeX2e
\usepackage[accepted]{tmlr}
\usepackage{amsmath,amsfonts,bm}

\def\eqref#1{equation~\ref{#1}}
\def\1{\bm{1}}

\DeclareMathAlphabet{\mathsfit}{\encodingdefault}{\sfdefault}{m}{sl}
\SetMathAlphabet{\mathsfit}{bold}{\encodingdefault}{\sfdefault}{bx}{n}

\usepackage{hyperref}
\usepackage{url}

\usepackage{amsmath}
\usepackage{amssymb}
\usepackage{algorithm}
\usepackage{algpseudocode}
\usepackage{graphicx}

\usepackage{booktabs}
\usepackage{multirow}
\usepackage{siunitx}

\usepackage{colortbl}
\usepackage{xcolor}
\usepackage{soul}

\usepackage{wrapfig}

\usepackage{longtable}
\usepackage{microtype}
\usepackage{enumitem}
\usepackage{tabularx}

\usepackage[compact]{titlesec}

\titlespacing*{\section}{0pt}{0.5ex}{0.5ex}
\titlespacing*{\subsection}{0pt}{0.5ex}{0.5ex}
\titlespacing*{\subsubsection}{0pt}{0.5ex}{0.5ex}
\titlespacing*{\paragraph}{0pt}{0pt}{0.5ex}

\newcommand{\vect}[1]{\mathbf{#1}}

\definecolor{softblue}{RGB}{235, 248, 255}
\definecolor{softorange}{RGB}{255, 247, 235}
\definecolor{impgreen}{RGB}{0, 165, 0}

\newcommand{\res}[2]{#1}
\newcommand{\noimp}[1]{#1}

\title{ODC: Orthogonal Drift Correction for Improved \\Text-to-Image Semantic Alignment at Inference}

\author{%
  \name Kasra Kamali, Basim Azam, Tuan Duong Trinh, Jean Honorio, Naveed Akhtar \\
  \email \{kasra.kamali, tuanduong.trinh\}@student.unimelb.edu.au \\
  \email \{basim.azam, jean.honorio, naveed.akhtar1\}@unimelb.edu.au \\
  \addr School of Computing and Information Systems, The University of Melbourne
}

\def\month{10}  % Insert correct month for camera-ready version
\def\year{2026} % Insert correct year for camera-ready version
\def\openreview{\url{https://openreview.net/forum?id=BOdxfPdtjE}} % Insert correct link to OpenReview for camera-ready version

\begin{document}

\maketitle

\begin{abstract}
Text-to-image models have achieved remarkable success in generating high-quality images from textual descriptions. However, they often struggle with ``semantic drift,'' where the generated output fails to precisely align with complex or nuanced text prompts. While recent approaches have attempted to address semantic errors regarding attribute binding or object presence, there remains a gap for a more holistic method that addresses these issues by directly refining the text embeddings of the initial user prompt. In this work, we introduce Orthogonal Drift Correction (ODC), an inference-time guidance method designed to mitigate semantic drift without requiring model retraining or additional user inputs. ODC guides image generation through a two-stage process. In the first stage, it identifies the semantic drift by evaluating the initially generated image against the user prompt in a shared vision-language embedding space. It then isolates the component of this drift vector that is orthogonal to the prompt's direction and translates it back into text via a vocabulary-based surrogate mechanism. In the second stage, it produces refined text conditioning for a second generation pass by feeding both the initial text embedding and the re-embedded drift representation into an adaptive rank-reduced concept removal module. Our experiments demonstrate that ODC consistently improves prompt-image alignment, increasing CLIP scores by up to 5.4\% over the baseline on the \textsc{Coal}Bench-300 and PartiPrompts benchmarks. Furthermore, on the T2I-CompBench complex subset, ODC improves the average 3-in-1 score by 4.5\% over the competing methods. As a plug-and-play module, ODC offers a practical and efficient method for improving the reliability of text-to-image models. Code is available at \url{https://github.com/kasrakamali/ODC}.
\end{abstract}

\section{Introduction}
\label{introduction}

Modern text-to-image (T2I) models have opened new creative possibilities, allowing people to generate compelling images that would have been impossible to generate just a few years ago. Models such as Stable Diffusion \citep{rombach2022high}, Imagen \citep{saharia2022photorealistic}, and DALL-E 2 \citep{ramesh2022hierarchical} have demonstrated an unprecedented ability to synthesize visually compelling and artistic images from natural language descriptions, unlocking wide-ranging applications across numerous domains.

However, despite these profound capabilities, current state-of-the-art T2I models frequently fall short in generating images that precisely align with the full semantic meaning and composition of their input text prompts. This lack of precise semantic alignment, a phenomenon we term ``semantic drift'', pertains to the frequent failure of these models to precisely adhere to user prompts. For instance, they often misinterpret complex compositional instructions, leading to incorrect attribute binding (e.g., wrong material on objects), catastrophic neglect \citep{chefer2023attend} (omitting subjects from the prompt), or a failure to adhere to specified spatial relationships. This alignment gap forces users into frustrating trial-and-error prompt engineering, a practice that underscores the limitations of such models in accurately interpreting semantic intent.

This difficulty largely stems from how T2I models process textual information. State-of-the-art T2I diffusion models employ a U-Net architecture with cross-attention layers \citep{rombach2022high}, which are crucial for fusing visual and textual features. While these layers encode rich semantic relations, the interaction can sometimes skew feature representations, causing semantic misalignment between the input and output. A variety of approaches have been proposed to address this problem. Fine-tuning methods like DreamBooth \citep{ruiz2023dreambooth} and CoMat \citep{jiang2024comat}, personalize models for specific subjects but can be computationally demanding. Architectural expansions like ControlNet \citep{zhang2023adding} and T2I-Adapter \citep{mou2024t2i} provide explicit spatial control through auxiliary inputs but require additional inputs and specialized training. Alternatively, a growing class of methods focuses on training-free, inference-time manipulation of attention maps or latent representations. Techniques such as Attend-and-Excite \citep{chefer2023attend}, CONFORM \citep{meral2024conform}, and ToMe \citep{hu2024token} directly intervene in the generation process to ensure robust semantic binding and alignment. While effective for specific tasks such as subject presence and attribute binding, they do not address the misalignment caused by imperfect initial text embeddings. This body of work reveals a clear gap for a method that can efficiently improve text-image alignment by directly correcting the text embedding to resolve complex compositional failures.

In this paper, we address this gap by introducing Orthogonal Drift Correction (ODC), a unique  inference-time technique that directly corrects the text conditioning vector to prevent semantic drift. Our approach is based on the central hypothesis that semantic drift originates from the components of the computed text embedding that are orthogonal to the prompt's true semantic direction. ODC guides the image generation process through a two-stage mechanism. First, it generates an initial candidate image. Then, it leverages the  shared embedding space of a pre-trained vision-language model, to identify the semantic error vector between the generated image and the prompt. Crucially, instead of using the entire error, it isolates its orthogonal component via vector rejection. This component, which intuitively  represents the ``off-topic'' deviation, is then removed  from the initial text embedding. The resulting purified embedding guides a second generation pass, producing a final image with substantially higher prompt-image alignment. Our contributions can be summarized as follows:
\begin{itemize}[leftmargin=*]
   \item We identify \textit{orthogonal semantic drift} in text embeddings as a significant source of prompt-image misalignment in text-to-image models, and propose a model-agnostic, training-free, inference-time method that corrects the text conditioning embeddings without modifying the diffusion process or requiring fine-tuning.
   \item We propose an adaptive rank-reduced concept removal strategy that refines the text conditioning by processing the initial text embedding alongside a re-embedded drift representation, obtained via a vocabulary-based surrogate mechanism.
   \item We introduce \textsc{Coal}Bench-300 (\textbf{CO}mpositional \textbf{AL}ignment Benchmark), 300 synthetically generated prompts curated to rigorously probe semantic drift and prompt-image alignment. To support our method, we curate a large-scale vocabulary of over 300,000 words and phrases, compiled from multiple sources and refined through several preprocessing steps, to effectively capture and mitigate semantic drift.
\end{itemize}

\section{Background \& Related Work}
\label{background}

We provide the necessary context for understanding our proposed method below.

\paragraph{Prompt-Image Alignment.} Prompt-image alignment refers to the faithfulness of a generated image to the full semantic meaning and compositional structure of its input text prompt. While the goal of Text-to-Image (T2I) models is to translate natural language into visually aligned results, achieving precise alignment in terms of the spatial composition, complex layouts, poses, shapes, and forms described in a prompt remains a significant challenge \citep{huang2023t2i,saharia2022photorealistic}. This often necessitates numerous trial-and-error cycles from users to achieve their desired output. Our method contributes to improving prompt-image alignment without requiring these cycles.

\paragraph{Text-to-Image Models.} The dominant architecture in modern T2I generation is the Latent Diffusion Model (LDM) \citep{Rombach2022LDM}. Unlike earlier diffusion models that operated in the high-dimensional pixel space \citep{ho2020denoising}, LDMs perform the computationally intensive denoising process in the compressed latent space of a pre-trained variational autoencoder (VAE). This significantly reduces computational overhead while maintaining high-fidelity output. 

The core of an LDM's denoising process is a U-Net architecture \citep{ronneberger2015u}, which comprises a series of downsampling and upsampling blocks with residual connections. Interspersed within this network are self-attention layers, which capture global spatial dependencies, and crucially, cross-attention layers. These cross-attention layers are the primary mechanism through which textual guidance is integrated into the image generation process. They allow the model to attend to different parts of the text prompt at each denoising step, conditioning the visual features being generated. More advanced models like Stable Diffusion XL (SDXL) \citep{podell2023sdxl} enhance this architecture with a larger U-Net and a dual text encoder setup to achieve superior performance.

\paragraph{Text Embeddings and Conditioning.} The journey from a text prompt to a guiding signal begins with tokenization, where the input string is converted into a sequence of discrete tokens. These tokens are then mapped to high-dimensional embedding vectors. This sequence of vectors is processed by a powerful text encoder, such as the CLIP text encoder \citep{radford2021learning} or a larger language model like T5 \citep{raffel2020exploring}, to produce the final conditioning embedding.

This text embedding conditions the denoising U-Net at multiple resolutions via the cross-attention mechanism. At each cross-attention layer, the intermediate visual features (the query) attend to the text embedding (the key and value). This process generates spatial attention maps that define a rich semantic relationship between image regions and prompt tokens, critically influencing the final image's composition. Our  method, intervenes at the very beginning of this pipeline. It is plugged into the process by modifying the conditioning embedding \textit{before} it is  passed to the U-Net, thereby correcting the guidance signal at its source.

\paragraph{Previous Attempts to Improve Alignment.} The challenge of prompt-image misalignment has inspired a wide variety of solutions. Prior work can be broadly categorized into three categories: prompt engineering, training-based methods, and inference-time guidance.

\textbf{Prompt Engineering:} At the most fundamental level, users engage in manual prompt engineering, iteratively refining text to achieve desired results. More structured approaches include prompt weighting, which allows users to amplify or attenuate the semantic influence of specific words, and the use of negative prompts. Negative prompts modify the unconditional embedding used in classifier-free guidance to explicitly steer the model away from unwanted concepts \citep{ho2022classifier}. The efficacy of these prompts depends heavily on the capacity of the text encoder to capture complex relationships. \citet{saharia2022photorealistic} demonstrated that scaling the frozen text encoder (e.g., T5-XXL) yields significantly larger improvements in alignment than scaling the diffusion model itself. Beyond manual refinement and architectural scaling, researchers have also explored optimizing the input space directly to represent specific concepts. Textual Inversion \citep{gal2022image} optimizes new pseudo-words within the embedding space of a frozen model to capture unique subjects or styles, acting as a bridge between the user's intent and the model's vocabulary.

\textbf{Training-Based Methods:} A powerful but resource-intensive approach involves fine-tuning the model. DreamBooth \citep{ruiz2023dreambooth} fine-tunes the entire diffusion model to bind a unique identifier with a specific subject, allowing for the generation of that subject in novel contexts with high fidelity. To mitigate the high computational cost of full fine-tuning, Low-Rank Adaptation (LoRA) \citep{hu2022lora} freezes pre-trained weights and injects trainable rank-decomposition matrices, enabling efficient adaptation to new tasks or styles with minimal parameter updates. While these methods focus on personalization, other works target general alignment failures such as concept ignorance. CoMat \citep{jiang2024comat} introduces an end-to-end fine-tuning strategy that leverages an image-to-text captioning model to minimize the discrepancy between the prompt and the generated image's caption, aligning the model to revisit ignored concepts and correct attribute mismapping.

\textbf{Inference-Time Guidance:} This category includes methods that, like ours, operate at inference time without modifying model weights. The foundational technique in this category is Classifier-Free Guidance (CFG) \citep{ho2022classifier}, which mixes conditional and unconditional score estimates to trade off sample diversity for higher text alignment. Building on this, \citet{chung2024cfg++} identified that standard CFG suffers from ``off-manifold" issues that harm editability and alignment, proposing CFG++ to correct the sampling trajectory to remain on the data manifold. Beyond sampling adjustments, several methods intervene specifically in the cross-attention layers or latent optimization during inference to correct semantic failures. \citet{hertz2022prompt} introduced Prompt-to-Prompt, which injects cross-attention maps from a source generation into a new generation to maintain structural layout while editing semantic content. To address ``catastrophic neglect'', Attend-and-Excite \citep{chefer2023attend} optimizes the latent at each timestep to ensure all subject tokens maximize their attention activations. Similarly, CONFORM \citep{meral2024conform} applies a contrastive objective during inference to segregate objects in attention maps while keeping attributes close to their subjects. Taking a different approach to binding, \citet{hu2024token} proposed Token Merging (ToMe), which merges relevant tokens into a composite token and iteratively updates this embedding during the early denoising steps using semantic binding and entropy losses. Finally, rather than optimizing the denoising path, InitNO \citep{guo2024initno} focuses on the initialization phase. It partitions the initial latent space into valid and invalid regions based on attention conflict scores and optimizes the initial noise vector to ensure it is predisposed toward semantically faithful generation before denoising begins.

Crucially, our work differs from existing inference-time approaches in its point of intervention. Whereas prior methods act by altering the internal dynamics of the U-Net denoising process, our approach is simpler and more direct: we refine the conditioning embeddings before diffusion begins. In contrast to training-based approaches, our method is entirely training-free, requiring no fine-tuning or additional optimization. Similarly, unlike attention-manipulation techniques, our intervention operates directly on the text embeddings, making it straightforward to implement and agnostic to the choice of U-Net architecture. By correcting semantic drift at its source, ODC provides a generalizable and principled way to improve the alignment and reliability of pre-trained text-to-image models.

\section{Orthogonal Drift Correction}
\label{method}

Our proposed method, Orthogonal Drift Correction (ODC), is an inference-time guidance technique designed to enhance the semantic alignment of text-to-image models. The central hypothesis is that suboptimal prompt alignment, or \textit{semantic drift}, stems from the components of the initial text embedding that are orthogonal to the prompt's primary semantic direction. ODC operates in a two-stage process. An initial image is generated and then evaluated against the prompt in a shared vision-language embedding space. The identified semantic error is then used to refine the initial text embedding for a second, more accurate generation pass.

\subsection{Notations}

Let $p_{\text{txt}}$ denote the input text prompt. The text-to-image model consists of two text encoders, a token-level encoder $\mathcal{E}_{\text{tok}}$ and a pooled encoder $\mathcal{E}_{\text{pool}}$, together with an image generator $\mathcal{G}$ (typically a U-Net implementing a reverse diffusion process). The standard generation process first encodes the text prompt as
\begin{equation}
    \mathbf{E}_p = \mathcal{E}_{\text{tok}}(p_{\text{txt}}), 
    \qquad
    \mathbf{e}_p = \mathcal{E}_{\text{pool}}(p_{\text{txt}}),
\end{equation}
where $\mathbf{E}_p \in \mathbb{R}^{L \times H}$ is the sequence of per-token embeddings and $\mathbf{e}_p \in \mathbb{R}^{H}$ is a pooled sentence-level vector. The image generator then produces an initial image $I$ conditioned on both representations:
\begin{equation}
    I = \mathcal{G}(\mathbf{E}_p, \mathbf{e}_p).
\end{equation}

To evaluate and correct for semantic drift, we employ a pre-trained vision-language model (VLM), denoted $\mathcal{E}_{\text{VLM}}$, which embeds both text and images into a shared multimodal space.

\subsection{The Orthogonal Drift Correction (ODC) Algorithm}

\begin{figure}[t]
    \centering
    \includegraphics[width=0.90\textwidth]{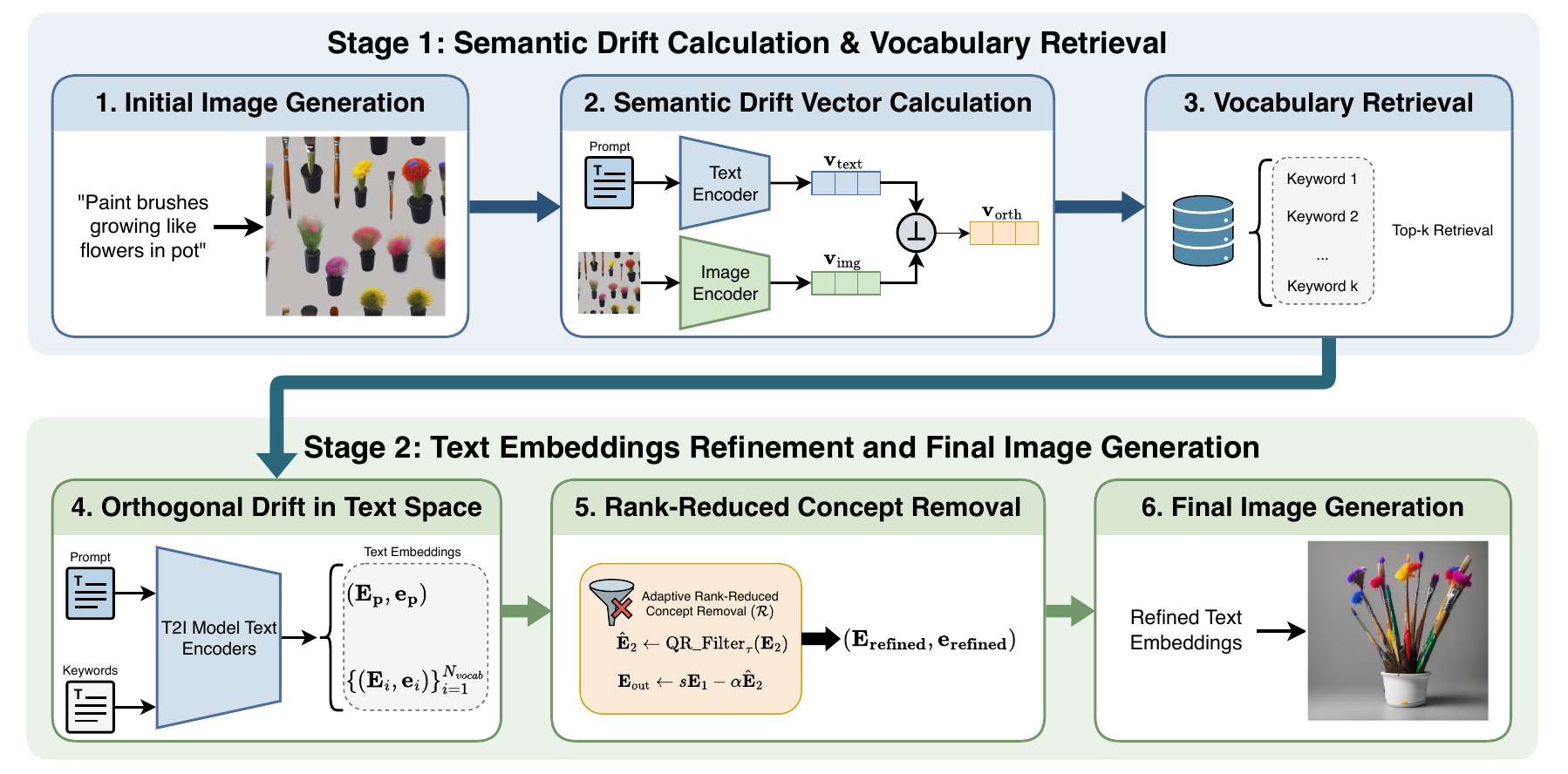}
    \vspace{\dimexpr-\intextsep\relax} 
    \caption{The workflow of Orthogonal Drift Correction (ODC). In the first stage, vocabulary items that best capture the orthogonal drift vector are retrieved. In the second stage, these vocabulary items are used in a concept removal operation, producing embeddings that better align with the user's text prompt and serve as input for the final image generation.}
    \label{fig:odc_workflow}
    \vspace{-10pt}
\end{figure}

The ODC algorithm consists of two main stages, comprising six sequential steps in total. These steps are described in detail below and summarized in Algorithm~\ref{alg:odc}. An overview of the process is illustrated in Figure~\ref{fig:odc_workflow}.

\paragraph{Step 1: Initial Image Generation.}
First, an initial image $I$ is generated using the standard process. The text prompt $p_{\text{txt}}$ is encoded by two complementary text encoders: one produces a sequence of per-token embeddings $\mathbf{E}_p$, while the other yields a pooled sentence-level representation $\mathbf{e}_p$. Both conditioning signals are then provided to the image generator $\mathcal{G}$ to synthesize the initial image $I$.

\paragraph{Step 2: Semantic Drift Vector Calculation.}
Next, we quantify the semantic discrepancy between the generated image $I$ and the original prompt $p_{\text{txt}}$. We use  $\mathcal{E}_{\text{VLM}}$ to project both onto a shared embedding space. The \textit{semantic drift vector}, $\vect{v}_{\text{drift}}$, captures the discrepancy between these two embeddings. This vector essentially encodes the semantic error, i.e., the direction and magnitude of the deviation of the image's content from the prompt's intent as perceived by the VLM.
\begin{align}
    \vect{v}_{\text{text}} = \mathcal{E}_{\text{VLM}}(p_{\text{txt}}), 
    \qquad
    \vect{v}_{\text{img}} = \mathcal{E}_{\text{VLM}}(I), 
    \qquad
    \vect{v}_{\text{drift}} = \vect{v}_{\text{img}} - \vect{v}_{\text{text}}.
\end{align}
\paragraph{Step 3: Orthogonal Drift Component Isolation and Vocabulary Retrieval.}
For $\vect{v}_{\text{drift}}$, we hypothesize that the most detrimental error components are those that introduce concepts extraneous to the prompt, which correspond to the direction orthogonal to the prompt's own embedding, $\vect{v}_{\text{text}}$. To isolate those components, we estimate the vector rejection of $\vect{v}_{\text{drift}}$ from $\vect{v}_{\text{text}}$. This is achieved by subtracting the projection of $\vect{v}_{\text{drift}}$ onto $\vect{v}_{\text{text}}$ from $\vect{v}_{\text{drift}}$ itself. The projection, $\vect{v}_{\text{proj}}$, is calculated as
\begin{equation}
    \vect{v}_{\text{proj}} = \left( \frac{\vect{v}_{\text{drift}} \cdot \vect{v}_{\text{text}}}{\|\vect{v}_{\text{text}}\|^2} \right) \vect{v}_{\text{text}}.
\end{equation}
The \textit{orthogonal drift vector}, $\vect{v}_{\text{orth}}$, is then the rejection:
\begin{equation}
    \vect{v}_{\text{orth}} = \vect{v}_{\text{drift}} - \vect{v}_{\text{proj}}.
\end{equation}
To obtain a discrete, textual surrogate for $\mathbf{v}_{\text{orth}}$, we retrieve candidate terms from a custom vocabulary. To this end, we construct a large-scale vocabulary retrieval dataset designed to cover both visually grounded concepts and general linguistic phrases. The dataset construction process involves several stages of aggregation and filtering. We provide details of the dataset and its construction in Section~\ref{vocab}. Each vocabulary item $w$ is associated with an embedding $\mathbf{e}_w$, and cosine similarities are computed as
\begin{equation}
    s(w) = \frac{\mathbf{v}_{\text{orth}} \cdot \mathbf{e}_w}{\|\mathbf{v}_{\text{orth}}\| \, \|\mathbf{e}_w\|}.
\end{equation}
The top-$k$ terms with the highest similarity scores are selected, forming a discrete approximation of $\mathbf{v}_{\text{orth}}$. We denote this retrieved set of tokens as $\mathcal{S}_{\text{orth}}$.

\paragraph{Step 4: Approximating Orthogonal Drift in  Text Encoder Space.}
The retrieved items $\mathcal{S}_{\text{orth}}$ are then re-embedded via the text encoders of the text-to-image model, yielding a per-token embedding matrix $\mathbf{E}_i$ and a pooled embedding vector $\mathbf{e}_i$ for each item $w_i \in \mathcal{S}_{\text{orth}}$. Since $\mathcal{S}_{\text{orth}}$ typically contains more than one item, we combine them using a softmax-weighted pooling to obtain equivalent representative embeddings:
\begin{align}
    \mathbf{E}_{\text{orth}} = \sum_{i} \beta_i \, \mathbf{E}_i, 
    \qquad
    \mathbf{e}_{\text{orth}} = \sum_{i} \beta_i \, \mathbf{e}_i,
\end{align}
where the weights $\beta_i$ are defined by the normalized cosine similarity scores:
\begin{equation}
    \beta_i = \frac{\exp(s(w_i))}{\sum_{j}\exp(s(w_j))}.
\end{equation}
This results in a pair $(\mathbf{E}_{\text{orth}}, \mathbf{e}_{\text{orth}})$ that represents the surrogate set in the embedding space of the text encoders of our T2I model. These representative embeddings, in addition to the original embeddings, are supplied to the concept removal module.

\paragraph{Step 5: Adaptive Rank-Reduced Concept Removal.}
To remove undesired semantic concepts from the text embeddings, we employ an adaptive rank-reduction approach that automatically identifies and retains the meaningful semantic components while filtering out noise. We denote this operator by $\mathcal{R}$, and the procedure of its computation is summarized in Algorithm~\ref{alg:adaptive_removal}.

Given embedding tensors $\mathbf{E}_1, \mathbf{E}_2 \in \mathbb{R}^{B \times L \times H}$ (batch size $B$, sequence length $L$, hidden dimension $H$), we first determine the semantic rank of $\mathbf{E}_2$ through QR decomposition. For each batch element:
\begin{equation}
    \mathbf{E}_2^{T} = \mathbf{Q}\mathbf{R},
\end{equation} 
where $\mathbf{Q}$ contains orthonormal basis vectors. We identify semantically meaningful components by selecting basis vectors whose relative magnitude exceeds $\zeta = 1\%$ of the primary component:
\begin{equation}
    k = \max\left\{ i : \frac{|r_{ii}|}{|r_{11}|} > \zeta \right\},
\end{equation}
where $r_{ii}$ are the diagonal elements of $\mathbf{R}$. We selected this threshold empirically to separate information from noise. Our analysis shows that while the majority of energy is concentrated in the first component, semantic distinctions are spread across several smaller components (ranks 3--15).

We then reconstruct a denoised version using only the top-$k$ basis vectors:
\begin{equation}
    \mathbf{E}_2^{\text{reduced}} = (\mathbf{E}_2 \mathbf{Q}_{:,:k}) \mathbf{Q}_{:,:k}^{T}.
\end{equation}
The concept removal operates on flattened representations to capture global semantic relationships. Let $\tilde{\mathbf{E}}_1 = \text{flatten}(\mathbf{E}_1)$ and $\tilde{\mathbf{E}}_2 = \text{flatten}(\mathbf{E}_2^{\text{reduced}})$. We compute:
\begin{equation}
    \tilde{\mathbf{E}}_{\text{out}} = \tilde{\mathbf{E}}_1 - \alpha \, \text{Rej}_{\tilde{\mathbf{E}}_1}(\tilde{\mathbf{E}}_2)
    = \tilde{\mathbf{E}}_1 \cdot \left(1 + \alpha \cdot \frac{\langle \tilde{\mathbf{E}}_1, \tilde{\mathbf{E}}_2 \rangle}{\|\tilde{\mathbf{E}}_1\|_2^2}\right) - \alpha \cdot \tilde{\mathbf{E}}_2,
\end{equation}
where $\alpha$ controls the removal strength. The output is reshaped to the original dimensions. This adaptive approach automatically adjusts to concept complexity, i.e., simple concepts require fewer ranks while complex concepts utilize more, achieving robust removal without over-filtering.

\paragraph{Step 6: Final Image Generation.}
Finally, a new image $I^*$ is generated using the refined conditioning embeddings $(\mathbf{E}_{\text{refined}}, \mathbf{e}_{\text{refined}})$ produced by the concept removal module. The image generator $\mathcal{G}$ then synthesizes:
\begin{equation}
    I^* = \mathcal{G}(\mathbf{E}_{\text{refined}}, \mathbf{e}_{\text{refined}}).
\end{equation}
By incorporating the correction derived from the surrogate embeddings, the refined conditioning better suppresses drift components while preserving the original semantic intent of the prompt.

\begin{figure}[t]
\centering
\begin{minipage}[t]{0.48\textwidth}
    \begin{algorithm}[H]
    \caption{Orthogonal Drift Correction (ODC)}
    \label{alg:odc}
    \begin{algorithmic}[1]
      \renewcommand{\baselinestretch}{1.16}\selectfont
        \State \textbf{Input:} Prompt $p_{\text{txt}}$, vocabulary $\mathcal{V}$
        \State \textbf{Output:} Final image $I^*$

        \State $(\mathbf{E}_p, \mathbf{e}_p) \gets (\mathcal{E}_{\text{tok}}(p_{\text{txt}}), \mathcal{E}_{\text{pool}}(p_{\text{txt}}))$
        \State $I \gets \mathcal{G}(\mathbf{E}_p, \mathbf{e}_p)$

        \State $\mathbf{v}_{\text{txt}} \gets \mathcal{E}_{\text{VLM}}(p_{\text{txt}})$
        \State $\mathbf{v}_{\text{img}} \gets \mathcal{E}_{\text{VLM}}(I)$
        \State $\mathbf{v}_{\text{drift}} \gets \mathbf{v}_{\text{img}} - \mathbf{v}_{\text{txt}}$
        \State $\mathbf{v}_{\text{proj}} \gets \frac{\mathbf{v}_{\text{drift}} \cdot \mathbf{v}_{\text{txt}}}{\|\mathbf{v}_{\text{txt}}\|^2} \,\mathbf{v}_{\text{txt}}$
        \State $\mathbf{v}_{\text{orth}} \gets \mathbf{v}_{\text{drift}} - \mathbf{v}_{\text{proj}}$

        \For{each $w \in \mathcal{V}$}
            \State $\mathbf{e}_w \gets \text{embedding}(w)$
            \State $s(w) \gets \frac{\mathbf{v}_{\text{orth}}\cdot \mathbf{e}_w}{\|\mathbf{v}_{\text{orth}}\|\|\mathbf{e}_w\|}$ 
        \EndFor
        \State $\mathcal{S}_{\text{orth}} \gets \text{Top-}k \text{ items by } s(w)$
        \State Re-embed $\mathcal{S}_{\text{orth}}$: $\{(\mathbf{E}_i, \mathbf{e}_i)\}_{w_i \in \mathcal{S}_{\text{orth}}}$ 
        \State Compute weights $\alpha_i \gets \frac{\exp(s(w_i))}{\sum_j \exp(s(w_j))}$
        \State $\mathbf{E}_{\text{orth}} \gets \sum_i \beta_i \mathbf{E}_i,\quad \mathbf{e}_{\text{orth}} \gets \sum_i \beta_i \mathbf{e}_i$

        \State $(\mathbf{E}_{\text{refined}}, \mathbf{e}_{\text{refined}}) 
    \gets \mathcal{R}\big((\mathbf{E}_p, \mathbf{e}_p), (\mathbf{E}_{\text{orth}}, \mathbf{e}_{\text{orth}})\big)$
        \State $I^* \gets \mathcal{G}(\mathbf{E}_{\text{refined}}, \mathbf{e}_{\text{refined}})$

        \State \textbf{return} $I^*$

    \end{algorithmic}
    \end{algorithm}
\end{minipage}\hfill
\begin{minipage}[t]{0.50\textwidth}
    \begin{algorithm}[H]
        \caption{Adaptive Rank-Reduced Concept Removal $\; (\mathcal{R})$}
        \label{alg:adaptive_removal}
        \begin{algorithmic}[1]
            \State \textbf{Input:} $\mathbf{E}_1, \mathbf{E}_2 \in \mathbb{R}^{B \times L \times H}$, $\mathbf{e}_1, \mathbf{e}_2 \in \mathbb{R}^{B \times H}$, $\alpha \in [0,1]$, $\zeta = 0.01$; bounds $k_{\min}=3$, $k_{\max}=15$         \State \textbf{Output:} $\mathbf{E}_{\text{out}} \in \mathbb{R}^{B \times L \times H}$, $\mathbf{e}_{\text{out}} \in \mathbb{R}^{B \times H}$         \For{$b = 1$ to $B$}
            \State $(\mathbf{Q}^{(b)}, \mathbf{R}^{(b)}) \gets \mathrm{QR\_decomposition}\!\big(\mathbf{E}_2^{(b)\top}\big)$         \State $\mathbf{r} \gets \mathrm{diag}\!\big(|\mathbf{R}^{(b)}|\big)$         \State $k \gets \max\{ i : r_i / r_1 > \zeta \}$         \State $k \gets \mathrm{clip}(k, k_{\min}, k_{\max})$         \State Let $\mathbf{Q}_k^{(b)}$ be the first $k$ columns of $\mathbf{Q}^{(b)}$         \State $\mathbf{E}_{2,\text{reduced}}^{(b)} \gets \big(\mathbf{E}_2^{(b)} \mathbf{Q}_k^{(b)}\big)\mathbf{Q}_k^{(b)\top}$         \State $\tilde{\mathbf{E}}_1 \gets \mathrm{flatten}(\mathbf{E}_1^{(b)})$         \State $\tilde{\mathbf{E}}_2 \gets \mathrm{flatten}(\mathbf{E}_{2,\text{reduced}}^{(b)})$         \State $\text{norm}_1 \gets \lVert \tilde{\mathbf{E}}_1 \rVert_2$         \State $s \gets 1 + \alpha \cdot \langle \tilde{\mathbf{E}}_1 / \text{norm}_1, \tilde{\mathbf{E}}_2 / \text{norm}_1 \rangle$         \State $\tilde{\mathbf{E}}_{\text{out}} \gets s \cdot \tilde{\mathbf{E}}_1 - \alpha \cdot \tilde{\mathbf{E}}_2$         \State $\mathbf{E}_{\text{out}}^{(b)} \gets \mathrm{reshape}(\tilde{\mathbf{E}}_{\text{out}}, [L, H])$         \EndFor
            
            \State $\text{norm}_e \gets \lVert \mathbf{e}_1 \rVert_2$         \State $s_e \gets 1 + \alpha \cdot \langle \mathbf{e}_1 / \text{norm}_e, \mathbf{e}_2 / \text{norm}_e \rangle$         \State $\mathbf{e}_{\text{out}} \gets s_e \cdot \mathbf{e}_1 - \alpha \cdot \mathbf{e}_2$         
            \State \Return $\mathbf{E}_{\text{out}}$, $\mathbf{e}_{\text{out}}$     \end{algorithmic}
    \end{algorithm}
\end{minipage}
\end{figure}

\section{Vocabulary Retrieval Dataset and COALBench-300}
\label{vocab_coalbench}

\subsection{Construction of the Vocabulary Retrieval Dataset}
\label{vocab}

To represent the isolated orthogonal drift vector $\mathbf{v}_{\text{orth}}$ using discrete textual surrogates, we constructed a vocabulary retrieval dataset of over 300,000 words and phrases designed to cover a diverse set of visually grounded and general linguistic concepts. The data was compiled from the following sources:

\begin{itemize}[leftmargin=*]
    \item \textbf{Open Images} \citep{OpenImages2}: Object class labels provided in the official release.
    \item \textbf{Visual Genome} \citep{krishna2017visual}: Object annotations, which supply a wide range of free-form object names contributed by annotators.
    \item \textbf{Wikipedia} \citep{wikipedia}: To augment the visual vocabulary with more general linguistic coverage, we parsed 1,000 articles extracted from English Wikipedia dataset (March 1, 2022 dump) using SpaCy's dependency parser \citep{spacy}. From each article, we extracted noun chunks, retaining those appearing at least three times across the corpus. This yielded a complementary set of multi-word expressions such as ``machine learning'' or ``climate change.''
\end{itemize}

In addition, we used a curated English word list (i.e., the \path{web2} dictionary via the \path{english-words} python package) to ensure broad lexical coverage \citep{english_words_pkg}.

Following data collection from these sources, we apply a comprehensive preprocessing phase to clean and refine the dataset. This procedure consists of three primary stages: normalization, heuristic filtering, and spelling correction.

\paragraph{Normalization.}
All candidate strings were lowercased, Unicode-normalized, stripped of non-alphanumeric symbols, and reduced to tokens separated by single spaces. This step removed morphological noise such as capitalization, punctuation, and diacritics.

\paragraph{Heuristic filtering.}
To eliminate uninformative entries, we applied a series of heuristic rules. First, we enforced length constraints, retaining only terms that contain 1 to 3 tokens and span between 3 and 30 characters. Next, we discarded strings dominated by digits or punctuation, as well as isolated stopword phrases (e.g., ``and'', ``the''). Finally, to ensure that short but meaningful concepts were not mistakenly removed by these length constraints, we implemented a whitelist to preserve essential terms such as ``tv'' and ``pc''.

\paragraph{Spelling correction.}
Because crowdsourced annotations introduce spelling inconsistencies, we employed the SymSpell algorithm \citep{symspell_alg} with a maximum edit distance of 2 to canonicalize misspellings. This reduced duplication from typographic errors.

This multi-stage process produced a broad and diverse vocabulary containing both concrete, visually grounded object names and higher-level conceptual phrases. In total, the final vocabulary contains approximately 323,000 items, which serves as the vocabulary retrieval dataset for our experiments.

\subsection{Construction of COALBench-300}
\label{coalbench}

In order to evaluate the semantic alignment between text-to-image models and user intent, we created \textsc{Coal}Bench-300 (\textbf{CO}mpositional \textbf{AL}ignment Benchmark), a curated collection of 300 imaginative prompts specifically designed to test challenging semantic alignment scenarios not emphasized in existing benchmarks. The goal was to establish a standardized set of inputs characterized by richness and diversity such that alignment failures would be clearly observable. To this end, we queried three distinct large language models: OpenAI o3 (\path{o3-2025-04-16}), Anthropic Claude Opus-4 (\path{claude-opus-4-20250514}), and Google Gemini 2.5 Pro (\path{gemini-2.5-pro}), with instructions to propose creative and highly visual text-to-image prompts. Each model produced a large pool of candidate prompts emphasizing diverse styles, scenarios, and semantic complexity.

From the generated pool, we manually reviewed and curated a final set of 300 prompts. This final collection of curated prompts provides a compact but challenging benchmark that allows us to systematically probe how well text-to-image models respect compositional semantics. Unlike generic caption corpora, these prompts emphasize stress-testing of alignment through imaginative and specific requests. The resulting dataset, alongside details of our filtering criteria and validation process is presented in Appendix~\ref{app:coalbench}.

\section{Experiments}
\label{experiments}

\subsection{Experimental Setup}

\paragraph{Models.}
To demonstrate the model-agnostic nature of ODC, we test it across Stable Diffusion XL (SDXL) \citep{podell2023sdxl} and FLUX.1 [schnell] \citep{BlackForestLabs2024Announcing}. We use the BLIP2-ITM-ViT-G model \citep{li2023blip} for all first-stage operations in ODC, namely, to compute the image and text embeddings for semantic drift estimation, to embed the vocabulary items, and to perform the vocabulary retrieval based on cosine similarity.

\paragraph{Inference Parameters.}
We used the default inference parameters recommended by the model authors for all experiments. This includes an image resolution of 1024$\times$1024 pixels, with SDXL running the Euler Discrete sampler (30 steps, CFG=7.5), and FLUX.1 [schnell] running the Flow Matching Euler Discrete sampler (4 steps). To ensure statistical robustness, we generated three images per prompt using fixed seeds across all methods and report the mean values for all metrics. Our method introduces a few inference-time hyperparameters. To determine their optimal values, we conducted a hyperparameter search on a randomly sampled, held-out 100-prompt subset of \textsc{Coal}Bench-300. Hyperparameters were selected once per model backbone and kept fixed across all evaluation benchmarks. The final selected configurations, along with ablation plots detailing performance across different parameter values, are provided in Appendix~\ref{app:hyperparameters}.

\paragraph{Benchmarks and Datasets.}
We evaluate our method using three distinct sets of prompts. First, we employ \textsc{Coal}Bench-300, introduced in Section \ref{coalbench}. To ensure a rigorous evaluation, the main quantitative results in Table~\ref{tab:main_results} are evaluated only on the 200 prompts held out from hyperparameter tuning, while other analyses in this paper use the complete set of 300 prompts. Second, we utilize PartiPrompts \citep{yu2022scaling} to probe imaginative and open-ended generation. For our experiments, we filter this dataset to retain only prompts containing ten words or more, ensuring that all evaluated prompts provide sufficient semantic grounding. Lastly, we employ T2I-CompBench (Complex) \citep{huang2025t2i}, a challenging benchmark subset that evaluates the generation of complex scenes involving multiple objects, attributes, and relationships.

\paragraph{Baselines.}
We compare ODC against several baselines to better understand its performance. We use the unmodified output from the base T2I models as our Vanilla reference. We also perform a CFG-Sweep, reporting results across different scales to compare against simply strengthening prompt guidance. Additionally, we evaluate a Negative Prompt baseline, which serves as a prompt-based alternative to our conditioning space correction. Specifically, we create a tailored negative prompt for each generation by joining the retrieved drift vocabulary items ($\mathcal{S}_{\text{orth}}$), instructing the model to avoid the identified spurious concepts. Finally, we compare against two specific inference-time guidance approaches, Attend-and-Excite \citep{chefer2023attend}, which refines the latent code during generation by optimizing cross-attention units to ``excite'' neglected subject tokens (adapted for SDXL; see Appendix~\ref{app:ae_details}), and ToMe \citep{hu2024token}, which merges object and attribute embeddings into composite tokens to unify their cross-attention maps via iterative optimization and entropy regularization.

\paragraph{Evaluation Metrics.}
We use established metrics for text-image alignment and computational efficiency. To measure semantic alignment between a prompt and its generated image, we rely on both CLIPScore (ViT-L/14) \citep{hessel2021clipscore}, a widely used metric for general semantic similarity, and BLIPScore (BLIP2-ITM-ViT-G) \citep{li2023blip}, which leverages the more advanced BLIP-2 vision-language model for finer-grained evaluations. Additionally, for the T2I-CompBench (Complex) benchmark, we report the ``3-in-1'' score alongside the individual component scores (BLIP-VQA, UniDet, and CLIPScore) to assess performance across attribute binding, spatial arrangement, and semantic alignment. Given that our method operates during inference, we measure its computational overhead in terms of both latency, defined as the average wall-clock time required to generate a single image, and GPU memory usage during generation.

\subsection{Quantitative Results}

To establish the efficacy of our method, we present a thorough quantitative evaluation. As shown in Tables~\ref{tab:main_results} and~\ref{tab:sota_comparison}, ODC consistently improves all baselines across semantic faithfulness metrics (CLIP and BLIP) on both Stable Diffusion XL and FLUX.1 [schnell] (statistical significance validated in Appendix~\ref{app:significance}). 

\begin{table}[t]
\centering
\caption{Main quantitative results on \textsc{Coal}Bench-300 and PartiPrompts benchmarks. We report alignment metrics across SDXL and FLUX.1 [schnell]. Best results are in \textbf{bold}. \colorbox{softblue}{Blue} and \colorbox{softorange}{Orange} rows highlight our methods. ODC BoT selects the best image from the two available choices---see text for details.}
\label{tab:main_results}

\renewcommand{\arraystretch}{1.2}
\setlength{\tabcolsep}{12pt} 

\resizebox{\linewidth}{!}{%
\begin{tabular}{lcccccccc}
\toprule
\multirow{3}{*}{\textbf{Method}} 
& \multicolumn{4}{c}{\textsc{Coal}Bench-300}
& \multicolumn{4}{c}{PartiPrompts} \\
\cmidrule(lr){2-5} \cmidrule(lr){6-9}
& \multicolumn{2}{c}{SDXL} 
& \multicolumn{2}{c}{FLUX.1 [schnell]}
& \multicolumn{2}{c}{SDXL}
& \multicolumn{2}{c}{FLUX.1 [schnell]} \\
\cmidrule(lr){2-3} \cmidrule(lr){4-5} \cmidrule(lr){6-7} \cmidrule(lr){8-9}
& CLIP $\uparrow$ & BLIP $\uparrow$ & CLIP $\uparrow$ & BLIP $\uparrow$ & CLIP $\uparrow$ & BLIP $\uparrow$ & CLIP $\uparrow$ & BLIP $\uparrow$ \\
\midrule
Vanilla & 30.3279 & 0.6078 & 29.4499 & 0.6076 & 30.8338 & 0.6030 & 31.2665 & 0.6051 \\
CFG-Sweep (CFG=5.0) & 30.0022 & 0.6070 & - & - & 30.4468 & 0.6021 & - & - \\
CFG-Sweep (CFG=12.0) & 30.4509 & 0.6083 & - & - & 31.1269 & 0.6040 & - & - \\
Negative Prompt & 30.3282 & 0.6078 & 29.4616 & 0.6074 & 30.8336 & 0.6030 & 31.1844 & 0.6051 \\
Negative Prompt BoT & 31.1898 & 0.6096 & 29.8912 & 0.6082 & 31.7436 & 0.6051 & 31.7811 & 0.6064 \\
\midrule
\rowcolor{softblue}
\textbf{ODC (Ours)} 
& \res{31.7976}{4.85} & \res{0.6121}{0.71} 
& \res{29.9181}{1.59} & \res{0.6084}{0.13}
& \res{32.5043}{5.42} & \res{0.6074}{0.73}
& \res{31.5468}{0.90} & \noimp{0.6051} \\

\rowcolor{softorange}
\textbf{ODC BoT (Ours)} 
& \textbf{\res{32.2907}{6.47}} & \textbf{\res{0.6127}{0.81}} 
& \textbf{\res{30.4549}{3.41}} & \textbf{\res{0.6097}{0.35}} 
& \textbf{\res{32.9911}{7.00}} & \textbf{\res{0.6082}{0.86}} 
& \textbf{\res{32.1538}{2.84}} & \textbf{\res{0.6067}{0.26}} \\
\bottomrule
\end{tabular}%
}
\vspace{-15pt}
\end{table}

\begin{table}[t]
\centering
\caption{Comparison with state-of-the-art. We report CLIP and BLIP scores for \textsc{Coal}Bench-300 and PartiPrompts benchmarks. For T2I-CompBench, we report the metrics on the complex subset, including the final 3-in-1 average score. Best results are in \textbf{bold}. All methods use SDXL as the baseline.}
\label{tab:sota_comparison}

\renewcommand{\arraystretch}{1.2}
\setlength{\tabcolsep}{12pt} 

\resizebox{\linewidth}{!}{%
\begin{tabular}{lcccccccc}
\toprule
\multirow{2}{*}{\textbf{Method}} 
& \multicolumn{2}{c}{\textsc{Coal}Bench-300} 
& \multicolumn{2}{c}{PartiPrompts}
& \multicolumn{4}{c}{T2I-CompBench (Complex)} \\
\cmidrule(lr){2-3} \cmidrule(lr){4-5} \cmidrule(lr){6-9}
& CLIP $\uparrow$ & BLIP $\uparrow$ & CLIP $\uparrow$ & BLIP $\uparrow$ & BLIP-VQA $\uparrow$ & UniDet $\uparrow$ & CLIP $\uparrow$ & 3-in-1 $\uparrow$ \\
\midrule
SDXL \citep{podell2023sdxl} & 30.305 & 0.608 & 30.834 & 0.603 & 0.442 & \textbf{0.025} & 0.293 & 0.3294 \\
A\&E \citep{chefer2023attend} & 29.542 & 0.608 & 30.15 & 0.604 & 0.436 & 0.023 & 0.290 & 0.3259 \\
ToMe \citep{hu2024token} & 30.397 & 0.608 & 30.681 & 0.603 & 0.468 & 0.022 & 0.292 & 0.3417 \\
\midrule 
\rowcolor{softblue}
\textbf{ODC (Ours)} 
& \textbf{\res{31.741}{4.74}} 
& \textbf{\res{0.612}{0.66}} 
& \textbf{\res{32.503}{5.41}} 
& \textbf{\res{0.607}{0.66}} 
& \textbf{\res{0.494}{11.76}} 
& 0.020 
& \textbf{\res{0.300}{2.39}} 
& \textbf{\res{0.3572}{8.44}} \\ 
\bottomrule
\end{tabular}%
}
\vspace{-15pt}
\end{table}

While ODC improves semantic faithfulness metrics, it can occasionally yield trade-offs in specific spatial metrics, such as UniDet (Table~\ref{tab:sota_comparison}). Although BLIP2-ITM-ViT-G is used both in our pipeline and the BLIP Score evaluation, the observed improvements are independently corroborated by CLIP Score, qualitative results, and human evaluation. Additionally, our VLM ablation study (Section~\ref{sec:vlm_ablation}) demonstrates that ODC yields consistent improvements across different VLM architectures, confirming that the method does not merely overfit to the BLIP2 embedding space. Since ODC's pipeline inherently produces both the initial baseline image (required for drift calculation) and the corrected variant, we can adopt a best-of-two (ODC BoT) strategy. Specifically, ODC BoT computes the CLIP Score for both the baseline and the ODC-corrected image, and returns the image with the higher score. This guarantees that the final output is never worse than the baseline in terms of CLIP alignment. To ensure a fair comparison, we also evaluate a Negative Prompt BoT baseline, which applies the same selection strategy to the Negative Prompt method (since it also requires an initial generation to retrieve drift vocabulary). As shown in Table~\ref{tab:main_results}, the Negative Prompt BoT baseline underperforms our method as well.

Furthermore, ODC preserves visual fidelity. We verify this using both an automated metric and human evaluation. For the automated evaluation, we measure the distributional shift between the original and corrected images using Kernel Inception Distance (KID) \citep{binkowski2018demystifying}. Near-zero KID scores (ranging from $-0.00152$ to $0.00093$) across SDXL and FLUX.1 confirm that the corrected distributions remain statistically indistinguishable from the baselines. Crucially, a human evaluation of visual fidelity found no statistically significant difference between baseline and ODC-corrected images (Binomial test $p=0.93$, Cohen's $h=0.13$), with a high tie rate of 58.0\%. This demonstrates that ODC surgically removes drift without shifting the generation out of the data manifold. See Appendix~\ref{app:kid} for details.

\begin{wraptable}{r}{0.45\textwidth}
    \vspace{\dimexpr-\intextsep-\baselineskip\relax} 
    \centering
    \caption{Efficiency analysis on SDXL.}
    \label{tab:efficiency_results}
    \renewcommand{\arraystretch}{1.25}
    \begin{tabular}{lcc}
    \toprule
    \textbf{Method} & Latency (s/img) & Memory (GB) \\
    \midrule
    Vanilla & 3.637 & 23.015 \\
    ODC     & 7.560 & 23.015 \\
    \bottomrule
    \end{tabular}
    \vspace{-10pt}
\end{wraptable}

As shown in Table \ref{tab:efficiency_results}, ODC introduces some latency overhead while leaving memory consumption unchanged. Compared to the vanilla baseline, latency increases from 3.637 s/img to 7.56 s/img, reflecting the additional correction pass. Importantly, GPU memory usage remains constant at 23.015 GB, making the approach just as accessible in practice as the baseline methods. Overall, these results demonstrate that the gains in semantic alignment come with predictable and manageable efficiency costs.

Ultimately, we validated our quantitative results via a human evaluation consisting of 600 blind pairwise comparisons. These were collected from 12 domain experts evaluating 50 randomly sampled prompts from \textsc{Coal}Bench-300. As shown in Table~\ref{tab:human_eval}, ODC is preferred in 61.8\% of decisive votes (95\% CI: [57.0\%, 66.3\%]) versus the baseline's 38.2\%. This represents a statistically significant improvement, as the CI strictly excludes 50\%. Furthermore, a 30.7\% tie rate indicates that ODC preserves fidelity in already well-aligned images rather than applying unnecessary alterations.

\subsection{Qualitative Results}

\begin{wraptable}{r}{0.45\textwidth}
\vspace{\dimexpr-\intextsep-\baselineskip\relax}
\centering
\caption{Human evaluation results. Win rates are computed over decisive votes (ties excluded).}
\label{tab:human_eval}
\begin{tabular}{lcc}
\toprule
\textbf{Metric} & \textbf{Value} & \textbf{95\% CI} \\
\midrule
ODC Win Rate   & 61.8\% & [57.0\%, 66.3\%] \\
Baseline Win Rate & 38.2\% & [33.7\%, 43.0\%] \\
\midrule
Tie Rate          & 30.7\% & --- \\
\bottomrule
\end{tabular}
\vspace{-10pt}
\end{wraptable}

\begin{figure}[t]
    \centering
    \includegraphics[width=0.95\textwidth]{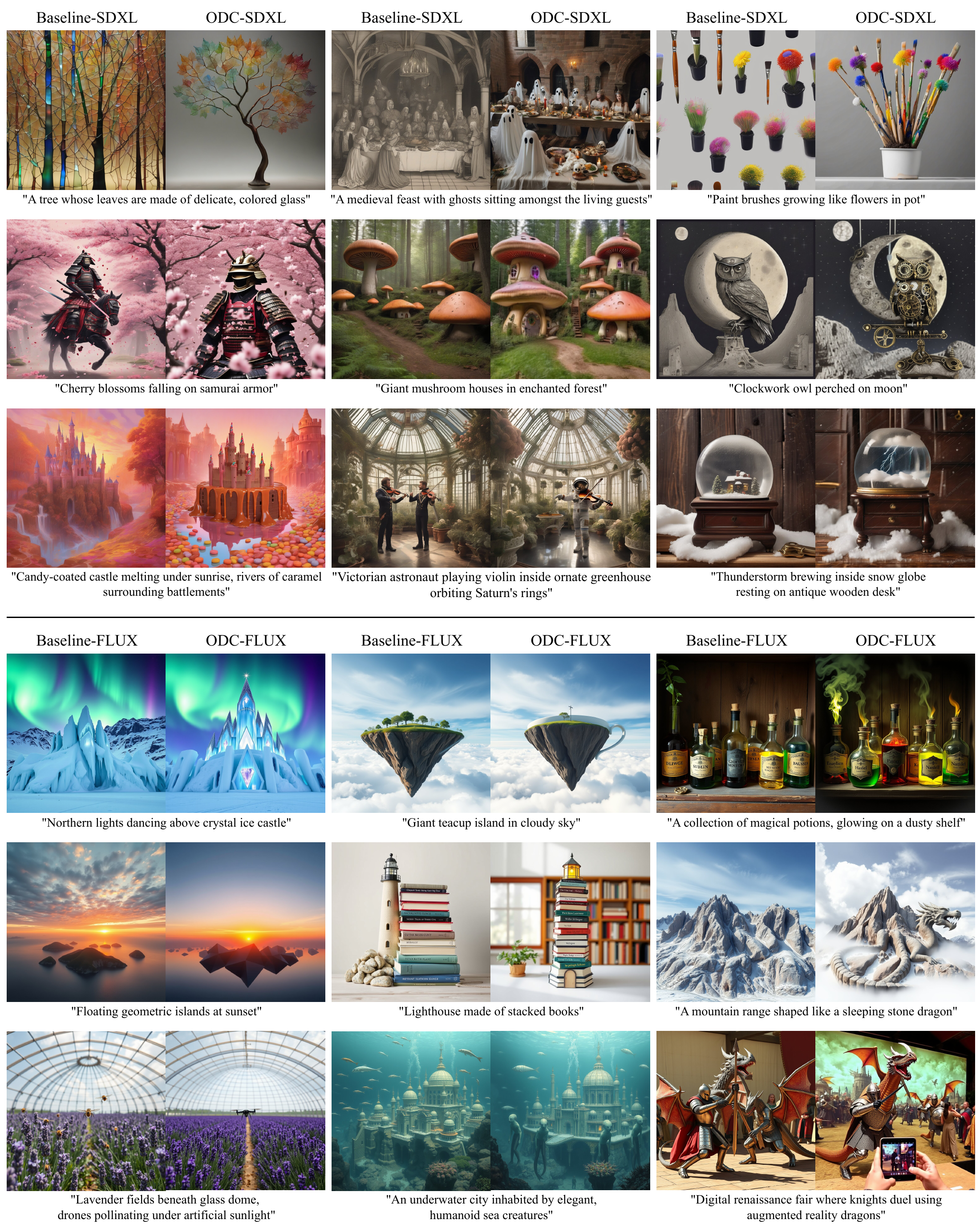}
    \vspace{\dimexpr-\intextsep\relax} 
    \caption{Qualitative comparison of images generated with and without Orthogonal Drift Correction (ODC). For each prompt, we show the output from the baseline model (left) and our ODC-corrected version (right).}
    \label{fig:qualitative_comparison}
    \vspace{-16pt}
\end{figure}

Visual inspection of the generated images provides the most intuitive evidence of ODC's effectiveness. In Figure~\ref{fig:qualitative_comparison}, we present a side-by-side comparison of images generated by vanilla SDXL and FLUX, alongside our ODC-corrected method for a variety of challenging prompts. We observe that ODC improves compositional accuracy and attribute binding, and leads to reduced semantic drift. The baseline often struggles with correct modifier-object relationships and maintaining fine-grained attributes, while ODC generates images that align more accurately with the user description. Furthermore, for prompts that are vulnerable to spurious additions (e.g., ``Victorian astronaut playing violin inside ornate greenhouse orbiting Saturn's rings''), the baseline model introduces unintended elements, such as additional violinists, that reduce alignment with the user prompt. In contrast, ODC constrains generations to the requested content and avoids unwanted additions, yielding outputs that more faithfully reflect the intended semantics. These qualitative examples demonstrate that by refining the initial text embedding, ODC provides a more reliable conditioning for the image generation process. Additional qualitative results are provided in Appendix~\ref{app:qualitative_appendix}. The list of vocabulary items ($\mathcal{S}_{\text{orth}}$) retrieved by ODC for images in Figure~\ref{fig:qualitative_comparison} are included in a detailed table in Appendix~\ref{app:drift_words}.

\begin{figure}[t]
    \centering
    \includegraphics[width=0.95\textwidth]{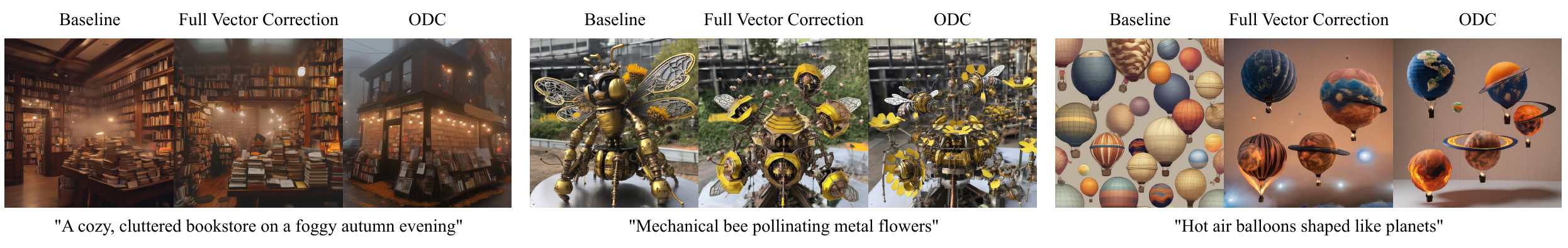}
    \vspace{\dimexpr-\intextsep\relax} 
    \caption{Qualitative ablation comparing baseline (SDXL), Full Vector Correction, and ODC on three prompts. Results show ODC corrects semantic discrepancies in the left and middle columns.}
    \label{fig:qualitative_ablation}
    \vspace{-5pt}
\end{figure}

\vspace{20pt}

\subsection{Ablation Studies}

\paragraph{Dependence on Vision-Language Model ($\mathcal{E}_{\text{VLM}}$).}
\label{sec:vlm_ablation}

To evaluate ODC's dependency on the VLM, we conducted an ablation study on \textsc{Coal}Bench-300 using three additional models of varying capacity: \texttt{CLIP ViT-L/14}, \texttt{SigLIP2 SO400M}, and \texttt{EVA-Giant Plus}. We also include our primary model, \texttt{BLIP2-ITM-ViT-G}, to serve as a reference point. For each VLM tested, all ODC first-stage operations are executed independently, with the respective model computing the semantic drift vector, embedding the vocabulary items, and performing the vocabulary retrieval to ensure the process operates entirely within its own embedding space. To ensure a rigorous comparison across these different embedding spaces, we calibrated ODC's removal strength ($\alpha$) on a held-out set of 50 prompts and evaluated performance on the remainder. As shown in Table~\ref{tab:vlm_ablation}, ODC generalizes across VLMs, with all models successfully improving baseline alignment. Among the evaluated VLMs, \texttt{BLIP2-ITM-ViT-G} achieves the highest overall alignment scores. \texttt{CLIP ViT-L/14} and \texttt{EVA-Giant Plus} outperform \texttt{SigLIP2}, with CLIP slightly outperforming the larger EVA on SDXL. We attribute this to architectural parity, as SDXL and FLUX.1 natively condition on the \texttt{CLIP ViT-L/14} text encoder, making its calculated drift vectors inherently better aligned with the generator's manifold. Ultimately, the strong performance of high-capacity models like \texttt{EVA-Giant Plus} and \texttt{BLIP2-ITM-ViT-G} confirms that such VLMs can effectively approximate this drift, ensuring ODC remains robust regardless of the chosen VLM.

\begin{table}[htbp]
  \centering
  \vspace{\dimexpr-\intextsep\relax} 
  \caption{Ablation Study on Vision-Language Model Dependency ($\mathcal{E}_{\text{VLM}}$).}
  \label{tab:vlm_ablation}
  \renewcommand{\arraystretch}{1.2}
  \setlength{\tabcolsep}{2pt} 
  \resizebox{\linewidth}{!}{%
  \begin{tabular}{@{}lcccccccccccccccc@{}}
    \toprule
    & \multicolumn{4}{c}{CLIP ViT-L/14} & \multicolumn{4}{c}{SigLIP2 SO400M} & \multicolumn{4}{c}{EVA-Giant Plus} & \multicolumn{4}{c}{BLIP2-ITM-ViT-G} \\
    \cmidrule(lr){2-5} \cmidrule(lr){6-9} \cmidrule(lr){10-13} \cmidrule(lr){14-17}
    & \multicolumn{2}{c}{SDXL} & \multicolumn{2}{c}{FLUX.1} & \multicolumn{2}{c}{SDXL} & \multicolumn{2}{c}{FLUX.1} & \multicolumn{2}{c}{SDXL} & \multicolumn{2}{c}{FLUX.1} & \multicolumn{2}{c}{SDXL} & \multicolumn{2}{c}{FLUX.1} \\
    \cmidrule(lr){2-3} \cmidrule(lr){4-5} \cmidrule(lr){6-7} \cmidrule(lr){8-9} \cmidrule(lr){10-11} \cmidrule(lr){12-13} \cmidrule(lr){14-15} \cmidrule(lr){16-17}
    \textbf{Method} & CLIP $\uparrow$ & BLIP $\uparrow$ & CLIP $\uparrow$ & BLIP $\uparrow$ & CLIP $\uparrow$ & BLIP $\uparrow$ & CLIP $\uparrow$ & BLIP $\uparrow$ & CLIP $\uparrow$ & BLIP $\uparrow$ & CLIP $\uparrow$ & BLIP $\uparrow$ & CLIP $\uparrow$ & BLIP $\uparrow$ & CLIP $\uparrow$ & BLIP $\uparrow$ \\
    \midrule
    Vanilla & 30.258 & 0.607 & 29.329 & 0.607 & 30.258 & 0.607 & 29.329 & 0.607 & 30.258 & 0.607 & 29.329 & 0.607 & 30.258 & 0.607 & 29.329 & 0.607 \\
    
    \rowcolor{softblue}
    \textbf{ODC} & 31.662 & 0.611 & 29.616 & 0.608 & 31.173 & 0.610 & 29.504 & 0.608 & 31.642 & 0.611 & 29.629 & 0.608 & 31.762 & 0.612 & 29.778 & 0.608 \\
    
    \rowcolor{softorange}
    \textbf{ODC BoT} & 32.106 & 0.612 & 30.105 & 0.609 & 31.559 & 0.611 & 30.141 & 0.609 & 32.126 & 0.612 & 30.354 & 0.610 & 32.206 & 0.612 & 30.278 & 0.610 \\
    \bottomrule
  \end{tabular}%
  }
  \vspace{-10pt}
\end{table}

\paragraph{The Importance of Orthogonal Correction.}

\begin{wraptable}{r}{0.45\textwidth}
    \vspace{\dimexpr-\intextsep-\baselineskip\relax} 
    \centering
    \caption{Quantitative ablation results for SDXL on \textsc{Coal}Bench-300.}
    \label{tab:ablation_orthogonal}
    \vspace{2pt}
    \renewcommand{\arraystretch}{1.25} 
    \setlength{\tabcolsep}{15pt} 
    
    \begin{tabular}{lc}
    \toprule
    \textbf{Method} & CLIP $\uparrow$ \\
    \midrule
    Vanilla & 30.305 \\
    Simple Concept Removal     & 30.970 \\
    Full Vector Correction     & 31.489 \\
    Orthogonal Drift Correction   & \textbf{31.741} \\
    \bottomrule
    \end{tabular}
\end{wraptable}

Our central claim is that isolating and removing the \textit{orthogonal} component of the semantic drift is key. To validate this, we compare ODC against two alternative correction strategies (Table~\ref{tab:ablation_orthogonal}). First, we test a ``Simple Concept Removal'' baseline, which directly subtracts the embeddings of the retrieved drift keywords from the prompt embeddings. While this improves upon the vanilla baseline, it substantially underperforms compared to the other approaches. Second, we test ``Full Vector Correction,'' which removes the entire semantic drift vector ($\mathbf{v}_{\text{drift}}$). Despite outperforming simple concept removal, it still underperforms our proposed ODC (Figure~\ref{fig:qualitative_ablation}), as removing the full vector inadvertently strips away useful on-axis semantic information. In contrast, ODC surgically removes only the irrelevant, orthogonal semantic contents, achieving the highest alignment.

\section{Discussion}
\label{sec:discussion}

While ODC operates in a training-free, plug-and-play manner, applying it to a new backbone requires a one-time calibration of its hyperparameters. This tuning burden is minimal and can be efficiently performed on a small held-out validation set or even calibrated intuitively based on a limited number of qualitative examples. Once these parameters are set for a given backbone, they remain fixed during evaluation.

The necessity of this per-backbone calibration stems from the fact that while ODC is model-agnostic, its magnitude of improvement is dependent on the specific model architecture's response to embedding space modifications. For instance, the alignment gains on FLUX.1 are smaller than on SDXL. We attribute this to FLUX.1's sensitivity to shifts in the embedding space. As demonstrated in our hyperparameter ablation (Appendix~\ref{app:hyperparameters}), FLUX.1 is prone to catastrophic over-correction at higher removal strengths (e.g., $\alpha=0.8$), necessitating a more conservative correction strength compared to SDXL. We hypothesize that this sensitivity may stem from how FLUX.1 processes and distributes token embeddings compared to SDXL, but leave a formal investigation of this to future work.

\section{Limitations}
\label{sec:limitations}

\begin{wrapfigure}{r}{0.35\textwidth}
  \vspace{\dimexpr-\intextsep\relax} 
  \centering
  \includegraphics[width=0.95\linewidth]{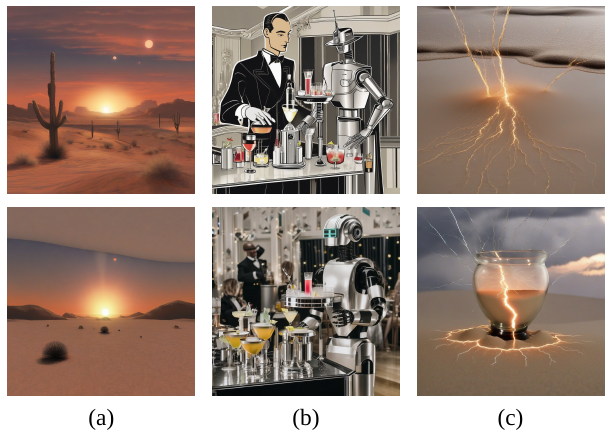}
  \vspace{-15pt}
  \caption{Representative failure cases: ODC (bottom) and Baseline (top).}
  \label{fig:failure_cases}
  \vspace{\dimexpr-\intextsep\relax}
\end{wrapfigure}

Despite its strong performance, ODC has some inherent limitations. First, its effectiveness is bound by the perceptual capabilities of the vision-language model (VLM) used in the first stage. If the VLM cannot see a specific semantic error, ODC  may not correct it. For instance, because VLMs struggle with precise counting, ODC might fail to correct numerical discrepancies, such as generating exactly ``A desert landscape with \underline{three suns} setting on the horizon'' (Figure~\ref{fig:failure_cases}(a)). Similarly, VLMs can miss nuanced artistic styles; if a baseline misses the intended style of a prompt like ``An \underline{Art Deco} robot butler serving cocktails at a party,'' the VLM may not recognize this drift, causing ODC to push the image towards realism rather than the desired style (Figure~\ref{fig:failure_cases}(b)). Furthermore, VLMs can misinterpret polysemous words, misguiding ODC. For example, with the prompt ``Lightning striking sand creating \underline{glass},'' the VLM might anchor onto the household object rather than the material, causing ODC to insert a literal drinking glass instead of a glassy formation (Figure~\ref{fig:failure_cases}(c)). These failures represent edge cases that are dependent on the interplay between the specific prompt and the generator, stemming from a combination of drift misestimation by the VLM and the generator's own susceptibility to hallucination. Because these factors are intertwined, simply swapping the base VLM may not reliably resolve all failures. However, future work could explore using task-specific evaluators (e.g., object-counting) to tune the hyperparameters for specific prompt types, or employing VLMs with stronger compositional reasoning to improve drift estimation. As VLM embeddings continue to improve, we expect ODC's robustness in these edge cases to advance as well.

Second, the two-stage correction process inherently increases the inference latency compared to a standard single pass, as shown in Table~\ref{tab:efficiency_results}. While the overhead from the embedding calculations themselves is negligible, the need for an initial generation pass represents a direct trade-off between computational cost and semantic drift correction. Nevertheless, considering that T2I generation is generally an offline task, this limitation is of minor nature. Moreover, our method can potentially reduce the time associated with manual trial and error process involved in prompt engineering considerably by providing better alignment of outputs to the prompts. 

\section{Conclusion}

In this paper, we addressed the critical problem of semantic drift in text-to-image models.  We  introduced Orthogonal Drift Correction (ODC), a training-free, inference-time method that guides the image generation through a two-stage process. It estimates the components of text embeddings causing the drift and mitigates them. Our extensive experiments demonstrate that ODC significantly enhances prompt-image alignment across multiple models and benchmarks. Without requiring any retraining or architectural changes, ODC provides a practical and generalizable solution for improving the alignment and reliability of existing pre-trained text-to-image models, making them more powerful creative tools.

\subsubsection*{Reproducibility Statement}
To ensure reproducibility, we provide detailed algorithmic descriptions of our method in Section~\ref{method}, along with hyperparameter configurations and experimental settings in Section~\ref{experiments}. Our proposed prompt dataset, \textsc{Coal}Bench-300, is provided in full in Appendix~\ref{app:coalbench}, and all other datasets used in our experiments are publicly available. All experiments are conducted using publicly available models and libraries. Our code is publicly available at \url{https://github.com/kasrakamali/ODC}.

\subsubsection*{Broader Impact Statement}

Improving prompt-image alignment improves reliability and reduces user frustration. Like any alignment technique, ODC makes models more faithful to their inputs, which could include harmful or inappropriate prompts if applied to base models with weak safety filters. Importantly, ODC does not expand the generative capabilities of the base model, but merely reduces semantic drift. To prevent misuse, ODC should be deployed in conjunction with standard content moderation and safety filtering mechanisms.

\subsubsection*{Acknowledgments}

Naveed Akhtar is a recipient of the Australian Research Council Discovery Early Career Researcher Award (project \#DE230101058) funded by the Australian Government. This research was supported by The University of Melbourne's Research Computing Services and the Petascale Campus Initiative, and by Melbourne Research Scholarship from The University of Melbourne.

\bibliography{main}
\bibliographystyle{tmlr}

\appendix

\setcounter{table}{0}
\renewcommand{\thetable}{A\arabic{table}}
\renewcommand{\theHtable}{A\arabic{table}}

\setcounter{figure}{0}
\renewcommand{\thefigure}{A\arabic{figure}}
\renewcommand{\theHfigure}{A\arabic{figure}}

\section{COALBench-300}
\label{app:coalbench}

Table~\ref{tab:dataset} presents the \textsc{Coal}Bench-300 dataset used in our experiments, consisting of 300 prompts generated as described in Section~\ref{coalbench}. To curate the final set from the initial generated prompts, we applied a filtering and validation process. First, we applied the following selection criteria: prompts were retained only if they (1) were highly imaginative and visually complex, requiring precise semantic adherence to execute correctly, and (2) contained clear descriptive constraints, such as specific visual properties, artistic styles, atmospheric conditions, or surreal combinations. Finally, to ensure independent validation, the curated list was reviewed and approved by a senior researcher, confirming that the selected prompts effectively serve as a stress test for compositional alignment. Furthermore, by providing the full list of 300 prompts in Table~\ref{tab:dataset}, we enable readers to directly assess and validate the benchmark.

\begin{longtable}{c p{0.35\textwidth} c p{0.35\textwidth}}
\caption{Full list of prompts in \textsc{Coal}Bench-300.} 
\label{tab:dataset} \\
\toprule
\textbf{\#} & \textbf{Prompt} & \textbf{\#} & \textbf{Prompt} \\
\midrule
\endfirsthead

\multicolumn{4}{c}%
{\tablename\ \thetable{} -- \textit{Continued from previous page}} \\
\toprule
\textbf{\#} & \textbf{Prompt} & \textbf{\#} & \textbf{Prompt} \\
\midrule
\endhead

\midrule
\multicolumn{4}{r}{\textit{Continued on next page}} \\
\endfoot

\bottomrule
\endlastfoot

1 & Neon cityscape reflected in rain puddles at midnight & 2 & Ancient library filled with floating glowing books \\
\addlinespace
3 & Steampunk octopus playing violin underwater & 4 & Cherry blossoms falling on samurai armor \\
\addlinespace
5 & Cyberpunk cat cafe with holographic menu displays & 6 & Dragon made entirely of autumn leaves \\
\addlinespace
7 & Astronaut planting flowers on Mars surface & 8 & Victorian greenhouse overflowing with bioluminescent plants \\
\addlinespace
9 & Robot chef preparing sushi in minimalist kitchen & 10 & Northern lights dancing above crystal ice castle \\
\addlinespace
11 & Abandoned amusement park reclaimed by nature & 12 & Phoenix rising from coffee cup steam \\
\addlinespace
13 & Underwater city inside giant glass bubble & 14 & Mechanical butterfly landing on child's outstretched hand \\
\addlinespace
15 & Storm clouds shaped like galloping horses & 16 & Art deco train station at golden hour \\
\addlinespace
17 & Jellyfish constellation floating through space & 18 & Medieval knight riding mechanical dragon \\
\addlinespace
19 & Sunflowers growing from piano keys & 20 & Time traveler's workshop filled with clocks \\
\addlinespace
21 & Magical forest where trees have doorways & 22 & Origami birds transforming into real ones \\
\addlinespace
23 & Vintage typewriter typing rainbow colored words & 24 & Giant teacup island in cloudy sky \\
\addlinespace
25 & Bioluminescent mushroom cave with crystal formations & 26 & Retro diner floating in outer space \\
\addlinespace
27 & Library ladder reaching into starry sky & 28 & Glass sculpture garden during thunderstorm \\
\addlinespace
29 & Mechanical whale swimming through clouds & 30 & Paint brushes growing like flowers in pot \\
\addlinespace
31 & Lighthouse made of stacked books & 32 & Geometric mountains with gradient sunset colors \\
\addlinespace
33 & Tiny fairy village inside terrarium & 34 & Steam locomotive made of flowers \\
\addlinespace
35 & Crystal skull containing miniature galaxy & 36 & Hot air balloons shaped like planets \\
\addlinespace
37 & Underwater ballet performed by seahorses & 38 & Gothic cathedral made of ice \\
\addlinespace
39 & Neon sign reflected in cyberpunk alley & 40 & Tree house connected by rope bridges \\
\addlinespace
41 & Desert oasis with mirror-like water & 42 & Mechanical heart with gears and flowers \\
\addlinespace
43 & Floating islands connected by waterfalls & 44 & Vintage camera capturing memories as butterflies \\
\addlinespace
45 & Snow globe containing miniature city & 46 & Giant mushroom houses in enchanted forest \\
\addlinespace
47 & Clock tower melting like Salvador Dali painting & 48 & Paper airplane leaving trail of stars \\
\addlinespace
49 & Underwater ruins with glowing hieroglyphs & 50 & Mechanical bee pollinating metal flowers \\
\addlinespace
51 & Aurora borealis inside crystal cave & 52 & Sailing ship made of autumn leaves \\
\addlinespace
53 & Robot gardener tending glass flowers & 54 & Mountain peak piercing through clouds \\
\addlinespace
55 & Neon Tokyo street in heavy rain & 56 & Butterfly wings made of stained glass \\
\addlinespace
57 & Ancient tree with doors to other worlds & 58 & Mechanical dragon breathing rainbow fire \\
\addlinespace
59 & Floating market on lily pad boats & 60 & Crystal palace reflecting sunset colors \\
\addlinespace
61 & Origami crane transforming into phoenix & 62 & Underwater volcano with lava lamp effect \\
\addlinespace
63 & Clockwork owl perched on moon & 64 & Garden maze viewed from above \\
\addlinespace
65 & Holographic whale swimming through city & 66 & Ice sculpture melting into butterflies \\
\addlinespace
67 & Vintage bicycle with flower basket & 68 & Portal opening in library wall \\
\addlinespace
69 & Mechanical spider weaving light web & 70 & Treehouse city in giant redwood forest \\
\addlinespace
71 & Glass elevator ascending through clouds & 72 & Firefly jar illuminating dark forest \\
\addlinespace
73 & Geometric deer made of neon lines & 74 & Floating tea party in sky \\
\addlinespace
75 & Crystal bridge over mirror lake & 76 & Mechanical hummingbird feeding on gear flowers \\
\addlinespace
77 & Northern lights reflected in frozen lake & 78 & Paper boat sailing on book pages \\
\addlinespace
79 & Bonsai tree with tiny glowing lanterns & 80 & Underwater castle made of coral \\
\addlinespace
81 & Steampunk hot air balloon city & 82 & Galaxy swirling in coffee cup \\
\addlinespace
83 & Ice phoenix rising from frozen waterfall & 84 & Mechanical butterfly collection in display case \\
\addlinespace
85 & Floating geometric islands at sunset & 86 & Glass greenhouse in snowy landscape \\
\addlinespace
87 & Lightning striking sand creating glass & 88 & Origami dragon breathing paper fire \\
\addlinespace
89 & Crystal cave with rainbow reflections & 90 & Vintage train crossing bridge at dawn \\
\addlinespace
91 & Mechanical whale breaching ocean surface & 92 & Fairy ring mushrooms glowing at twilight \\
\addlinespace
93 & Neon arcade in rain-soaked alley & 94 & Tree growing through abandoned car \\
\addlinespace
95 & Holographic fish swimming through air & 96 & Ice palace under aurora borealis \\
\addlinespace
97 & Steampunk telescope viewing distant galaxies & 98 & Garden path made of piano keys \\
\addlinespace
99 & Crystal butterfly landing on metal flower & 100 & Lighthouse beam cutting through fog \\
\addlinespace
101 & A bioluminescent jellyfish floating through a neon-lit cyberpunk city & 102 & A lone knight overlooking a valley of giant, glowing mushrooms \\
\addlinespace
103 & A crystal skull filled with a swirling, miniature galaxy & 104 & A serene, minimalist Japanese garden in the pouring rain \\
\addlinespace
105 & An ancient Roman library on a terraformed, dusty Mars & 106 & A tiny fairy house built inside a broken teapot \\
\addlinespace
107 & A grand piano with keys melting like liquid Dali-style & 108 & A cozy, cluttered bookstore on a foggy autumn evening \\
\addlinespace
109 & A Viking longship sailing on a sea of clouds & 110 & A stained glass window depicting a cosmic horror entity \\
\addlinespace
111 & A sleek, organic starship gliding silently past a vibrant nebula & 112 & A grizzly bear made of constellations in the night sky \\
\addlinespace
113 & A forgotten, overgrown greenhouse filled with strange, glowing plants & 114 & A city skyline at sunset, rendered in 8-bit pixel art \\
\addlinespace
115 & A steampunk inventor's workshop, cluttered with brass and clockwork & 116 & A macro photograph of a single dewdrop on a spiderweb \\
\addlinespace
117 & A majestic castle made of obsidian and jagged lightning & 118 & The concept of "silence" visualized as a physical object \\
\addlinespace
119 & An Art Deco robot butler serving cocktails at a party & 120 & A floating market on a serene, otherworldly alien canal \\
\addlinespace
121 & A portrait of a queen, made entirely from pressed flowers & 122 & A paper-craft, origami world with a flowing paper river \\
\addlinespace
123 & A cat made of glass, filled with swirling, colored smoke & 124 & A haunted forest where the trees have screaming human faces \\
\addlinespace
125 & Impossible Escher-like architecture carved from a single, giant tree & 126 & A gourmet hamburger made entirely of sparkling, edible gemstones \\
\addlinespace
127 & A desert landscape with three suns setting on the horizon & 128 & An ancient, moss-covered golem sleeping in a sunlit forest \\
\addlinespace
129 & A city street scene painted in the style of Van Gogh & 130 & A lone, red telephone box in the middle of nowhere \\
\addlinespace
131 & A mechanical heart with glowing tubes and intricate gears & 132 & A tranquil koi pond that reflects a starry night sky \\
\addlinespace
133 & An astronaut discovering an ancient alien ruin on the moon & 134 & A single, perfect rose encased in a block of ice \\
\addlinespace
135 & An island that is the back of a giant, swimming turtle & 136 & A shadowy figure standing at the end of a long pier \\
\addlinespace
137 & A black and white photograph of a 1940s detective's office & 138 & A city built inside a colossal, hollowed-out crystal cavern \\
\addlinespace
139 & A samurai warrior with glowing, cybernetic neon tattoos & 140 & A massive, ancient library where the books float between shelves \\
\addlinespace
141 & A cup of coffee with a swirling galaxy in it & 142 & A child releasing a kite that is actually a dragon \\
\addlinespace
143 & A detailed anatomical drawing of a mythical griffin & 144 & A vaporwave sculpture garden with pink and blue neon lights \\
\addlinespace
145 & A doorway to another dimension opening in a brick wall & 146 & A city street dissolving into musical notes in the air \\
\addlinespace
147 & A lone lighthouse battered by a furious, cosmic storm & 148 & A portrait of "joy" as an explosion of vibrant colors \\
\addlinespace
149 & A medieval feast with ghosts sitting amongst the living guests & 150 & An underwater city inhabited by elegant, humanoid sea creatures \\
\addlinespace
151 & A tree whose leaves are made of delicate, colored glass & 152 & A futuristic high-speed train blurring through a bamboo forest \\
\addlinespace
153 & A sad clown looking at his reflection in a puddle & 154 & A cinematic shot of a dinosaur roaring in a modern city \\
\addlinespace
155 & A map of a fantasy world burned onto old leather & 156 & An ornate, Venetian mask with intricate filigree and feathers \\
\addlinespace
157 & A car rusting in a field of blooming sunflowers & 158 & A mountain range shaped like a sleeping stone dragon \\
\addlinespace
159 & A wizard's spellbook lying open, glowing with powerful magic & 160 & A skeleton in a spacesuit, sitting on a desolate asteroid \\
\addlinespace
161 & A beautiful landscape painting done on a crumbling concrete wall & 162 & A ceramic mosaic depicting the birth of a star \\
\addlinespace
163 & A clock tower where the numbers are flying away & 164 & An elven city built seamlessly into giant, living trees \\
\addlinespace
165 & A low-poly render of a wolf howling at the moon & 166 & A secret garden hidden behind a waterfall \\
\addlinespace
167 & A single, burning feather falling from a dark sky & 168 & A long exposure photograph of highway lights at night \\
\addlinespace
169 & A palace made entirely of ice, glistening in the sun & 170 & A chessboard with living pieces preparing for battle \\
\addlinespace
171 & A surreal landscape where the ground is a reflective mirror & 172 & A powerful phoenix rising from ashes, cinematic fantasy art \\
\addlinespace
173 & A ukiyo-e woodblock print of a busy Tokyo intersection & 174 & An old, forgotten god statue entangled in jungle vines \\
\addlinespace
175 & A gentle robot tending to a garden of metal flowers & 176 & An underwater submarine discovering the lost city of Atlantis \\
\addlinespace
177 & The solar system re-imagined as a delicate, hanging mobile & 178 & A tiny mouse wearing intricate, handcrafted medieval armor \\
\addlinespace
179 & A world where the oceans are in the sky & 180 & A close-up of a dragon's eye, reflecting a battle \\
\addlinespace
181 & A hyper-realistic painting of a spilled glass of milk & 182 & A towering, complex sandcastle about to be hit by a wave \\
\addlinespace
183 & A lush, tropical beach inside a giant, open clam shell & 184 & A synthwave-style sunset over a retro-futuristic highway \\
\addlinespace
185 & A collection of magical potions, glowing on a dusty shelf & 186 & A double exposure portrait of a woman and a forest \\
\addlinespace
187 & A pirate ship sailing through a massive electrical storm & 188 & A whimsical town where all the buildings are books \\
\addlinespace
189 & A soldier's helmet from the future, cracked and abandoned & 190 & A tree of life connecting the earth and the cosmos \\
\addlinespace
191 & A blueprint-style technical drawing of a fantasy airship & 192 & A labyrinth made of tall, perfectly trimmed hedges \\
\addlinespace
193 & An abstract sculpture representing the sound of laughter & 194 & A street market on a space station, full of aliens \\
\addlinespace
195 & A beautiful, haunting portrait of a porcelain doll & 196 & The world balanced on the tip of a giant's finger \\
\addlinespace
197 & A polar bear sleeping under the vibrant aurora borealis & 198 & A vintage 1950s car flying through the clouds \\
\addlinespace
199 & A stone staircase spiraling down into the deep, dark ocean & 200 & A single, glowing key lying on an empty city street \\
\addlinespace
201 & Neon samurai meditating under cherry blossoms, cyberpunk city distant horizon & 202 & Golden dragon soaring above tempestuous sea during crimson twilight storm \\
\addlinespace
203 & Victorian astronaut playing violin inside ornate greenhouse orbiting Saturn's rings & 204 & Gigantic marble statues emerging from desert dunes beneath aurora sky \\
\addlinespace
205 & Steam-powered whale floating over steampunk city releasing glowing lanterns peacefully & 206 & Surreal library labyrinth with endless staircases and levitating luminous books \\
\addlinespace
207 & Winter fox spirits dancing around ancient shrine under full moon & 208 & Retro-futuristic diner on Mars serving milkshakes to robot cowboys today \\
\addlinespace
209 & Bioluminescent jungle river reflecting twin moons, explorers paddling crystal canoes & 210 & Minimalist ink painting depicting thunderstorm over solitary mountain tea house \\
\addlinespace
211 & Ancient turtle island carrying cherry forest drifting across tranquil ocean & 212 & Post-apocalyptic fashion runway among ruins, models wearing repurposed metallic couture \\
\addlinespace
213 & Glass butterflies emerging from cracked sculpture in abandoned classical gallery & 214 & Cosmic carnival Ferris wheel spinning galaxies instead of passenger gondolas \\
\addlinespace
215 & Medieval alchemist laboratory illuminated by swirling emerald and violet potions & 216 & Peaceful Zen garden floating among clouds, koi fish swimming air \\
\addlinespace
217 & Art deco submarine cruising crystal Arctic waters beneath shimmering aurora & 218 & Gothic cathedral built from bones, candles flickering with blue flames \\
\addlinespace
219 & Whimsical cat orchestra performing symphony inside giant teacup on meadow & 220 & Autumn maple leaves transforming into origami cranes mid gentle breeze \\
\addlinespace
221 & Futuristic Venice with glass canals and levitating gondolas at sunset & 222 & Alien botanist tending glowing succulents within transparent asteroid greenhouse sphere \\
\addlinespace
223 & Samurai mech duel atop skyscraper during torrential rain and lightning & 224 & Childhood memories visualized as floating polaroid cubes above grassy field \\
\addlinespace
225 & Mushroom village illuminated by fireflies, tiny gnomes celebrating midsummer feast & 226 & Surreal chessboard landscape where pieces are skyscrapers casting elongated shadows \\
\addlinespace
227 & Golden hour lighthouse standing on clouds guiding paper boats home & 228 & Cybernetic phoenix reborn from neon flames above dystopian alleyway rooftops \\
\addlinespace
229 & Mythical underwater library with coral shelves and manta ray librarians & 230 & Celtic wolf warrior howling beside rune-carved stone under starry sky \\
\addlinespace
231 & Industrial robots painting classical renaissance portrait inside abandoned factory hall & 232 & Pastel vaporwave beach scene with dolphins and pixelated palm trees \\
\addlinespace
233 & Ice cream planet orbiting candy stars within childlike cosmic dreamscape & 234 & Ethereal ballerina composed of smoke gracefully twirling through candlelit ballroom \\
\addlinespace
235 & Ancient astronauts carved into mountain, projections igniting during equinox ceremony & 236 & Monochrome noir city where rain turns neon once striking pavement \\
\addlinespace
237 & Giant panda knight wearing bamboo armor charging across sunflower plains & 238 & Sailing ship overhead sky ocean, clouds shaped like wandering whales \\
\addlinespace
239 & Magnificent art nouveau dragonfly airship docking at crystal palace port & 240 & Crimson haired witch brewing constellation soup in antique cast-iron cauldron \\
\addlinespace
241 & Time travelers picnic within looping paradox meadow filled with suns & 242 & Dreamy pastel clouds dripping like paint over sleepy coastal village \\
\addlinespace
243 & Crystal caverns resonating music, stalactites functioning as grand pipe organ & 244 & Pharaoh cyborg sitting upon holographic throne inside starship pyramid command \\
\addlinespace
245 & Origami dragons battling above waterfall, paper scraps becoming swirling mist & 246 & Intergalactic farmers harvesting stardust using laser scythes and hover wagons \\
\addlinespace
247 & Snowy owl postman delivering letters between lonely mountain observatories nightly & 248 & Renaissance painting fused with glitch art, saints pixelated yet divine \\
\addlinespace
249 & Lonely astronaut gardening roses inside cracked helmet on barren exoplanet & 250 & Midsummer carnival mirrored on lake surface, reflections showing alternate dimension \\
\addlinespace
251 & Celestial koi swimming through starry void forming luminous cosmic patterns & 252 & Gigabyte forest where data streams cascade like waterfalls between trees \\
\addlinespace
253 & Clockwork hummingbirds pollinating gears within enormous steampunk garden mechanism today & 254 & Futuristic samurai riding holographic horse across neon rice paddies night \\
\addlinespace
255 & Survivalists camping on colossal lily pads floating down moonlit river & 256 & Hyperrealistic microphotography style city built upon a dandelion seed closeup \\
\addlinespace
257 & Norwegian fjord village illuminated solely by bioluminescent algae and stars & 258 & Astral projection commuters riding light beams between futuristic office towers \\
\addlinespace
259 & Fairytale bakery selling clouds, customers bottling rainbows for frosting today & 260 & Desert caravan of holographic camels crossing dunes under binary suns \\
\addlinespace
261 & Spaceship graveyard turned skatepark, teenagers grinding along rusted hulls today & 262 & Fantasy opera house shaped like seashell, audience wearing pearl masks \\
\addlinespace
263 & Interactive graffiti walls repaint themselves according to passerby's emotional aura & 264 & Robot monk studying lotus algorithm amidst tranquil chrome bamboo grove \\
\addlinespace
265 & Abandoned amusement park reclaimed by jungle, rollercoaster entwined with vines & 266 & Crystalline wolf pack running across frozen lake under kaleidoscope aurora \\
\addlinespace
267 & Pocket watch melting into river, time flowing downstream with salmon & 268 & Digital renaissance fair where knights duel using augmented reality dragons \\
\addlinespace
269 & Samurai standing in field of wind turbines during golden sunrise & 270 & Galactic mail carrier surfing solar winds to deliver interstellar postcards \\
\addlinespace
271 & Carousel horses awakening midnight, galloping across city streets leaving sparks & 272 & Northern lights shaped like dragons over mirrorlike fjord with rowboat \\
\addlinespace
273 & Cyberpunk mermaid lounging on neon coral reef playing electric harp & 274 & Snow globe metropolis shaking, skyscraper flakes swirling around central park \\
\addlinespace
275 & Mystical doorway inside tree trunk leading to floating candlelit staircase & 276 & Marble chess pieces crying tears, puddles reflecting parallel universe conflicts \\
\addlinespace
277 & Steampunk octopus captain steering dirigible through thunderous cloud ocean today & 278 & Silk road marketplace set on asteroid, merchants trading gravity spices \\
\addlinespace
279 & Lone monk meditating atop newspaper stack skyscraper in bustling metropolis & 280 & Candy-coated castle melting under sunrise, rivers of caramel surrounding battlements \\
\addlinespace
281 & Moonlit bamboo forest where shadows of past lives perform theatre & 282 & Quantum computer envisioned as cathedral, qubits glowing like stained glass \\
\addlinespace
283 & Vibrant coral reef mirrored above water, sky filled with fish & 284 & Ghostly steam train crossing galaxy, passengers silhouettes made from stardust \\
\addlinespace
285 & Medieval tavern hosting extraterrestrial travelers swapping holographic treasure maps tonight & 286 & Arcade cabinet forest, joysticks sprouting like mushrooms among pixel leaves \\
\addlinespace
287 & Viking ship sailing across clouds towards colossal floating mead hall & 288 & Haunted typewriter printing poems by itself inside candlelit attic loft \\
\addlinespace
289 & Interdimensional bus stop where travelers trade memories for tickets home & 290 & Mid-century modern house perched atop giant redwood, overlooking fog valley \\
\addlinespace
291 & Lavender fields beneath glass dome, drones pollinating under artificial sunlight & 292 & Ancient rune compass projecting holographic maps above weathered pirate desk \\
\addlinespace
293 & Swan boat gliding through clouds, passengers sipping tea with angels & 294 & Neo-Tokyo skyline reflected in raindrop on cybernetic geisha's finger nail \\
\addlinespace
295 & Prismatic desert canyon walls refracting sunlight into rainbow patterned sand & 296 & Clocktower orchard where apples tick loudly with winding golden gears \\
\addlinespace
297 & Giraffe skyscrapers eating cloud leaves in whimsical metropolitan savanna today & 298 & Secluded monastery balanced atop waterfall edge within verdant hidden valley \\
\addlinespace
299 & Elegant cyborg ballerinas rehearsing in abandoned warehouse filled with mirrors & 300 & Thunderstorm brewing inside snow globe resting on antique wooden desk \\

\end{longtable}

\clearpage

\section{Hyperparameter Ablation}
\label{app:hyperparameters}

\begin{figure}[t]
  \centering
  \includegraphics[width=\textwidth]{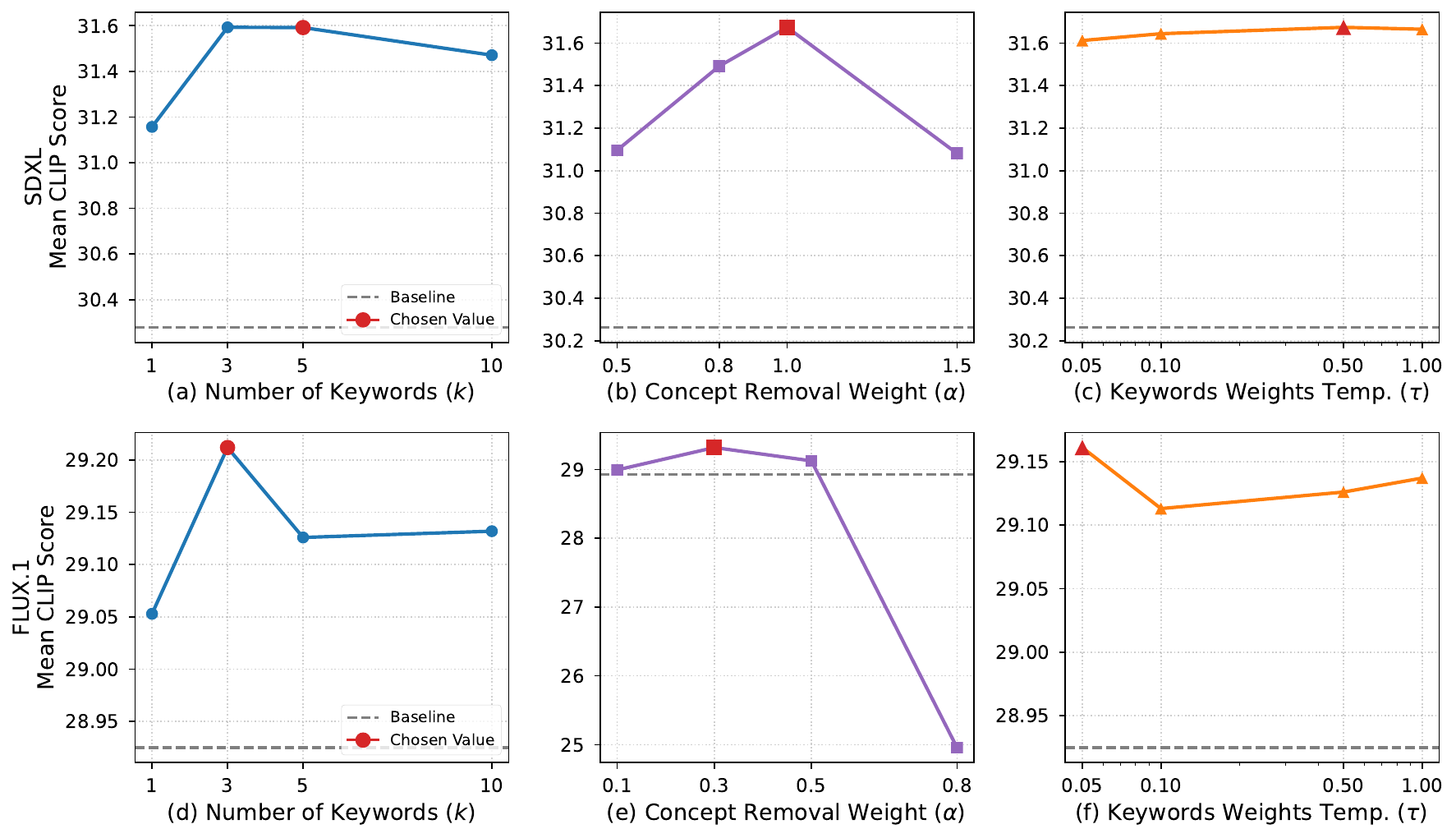}
  \caption{Hyperparameter ablation study on a 100-prompt validation subset of \textsc{Coal}Bench-300. Top row: SDXL. Bottom row: FLUX.1. The dashed line represents the unmodified baseline. Red markers indicate the hyperparameters chosen for our pipeline.}
  \label{fig:hyperparameters}
\end{figure}

\begin{figure}[htbp]
  \centering
  \includegraphics[width=\textwidth]{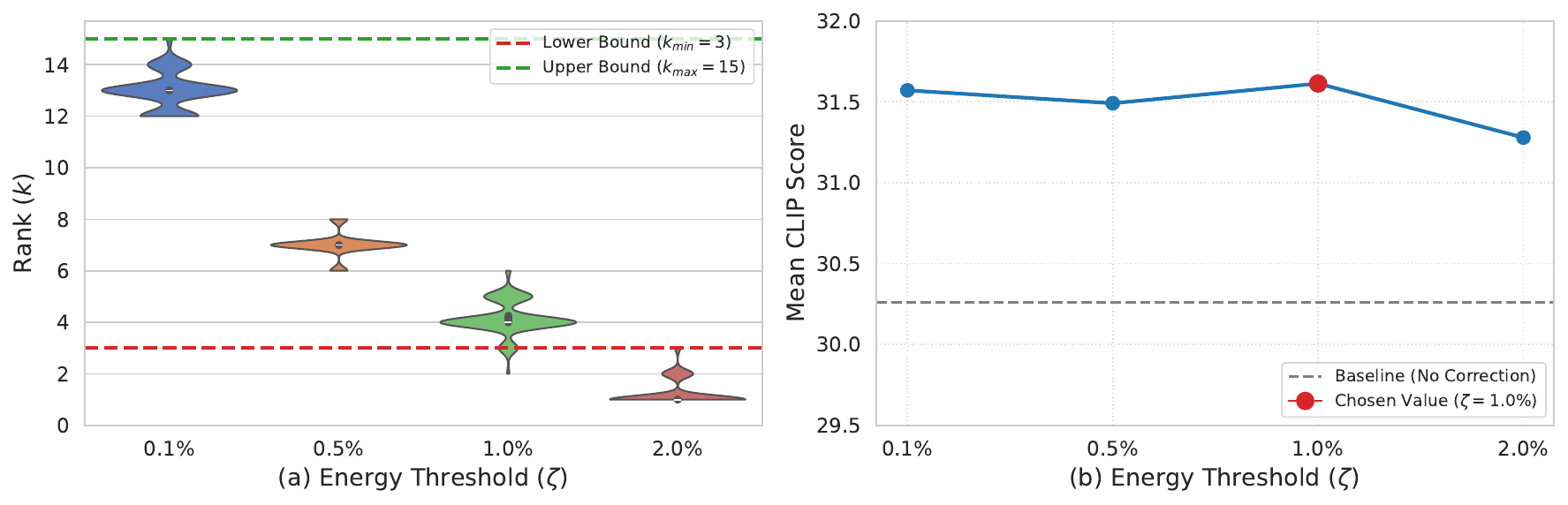}
  \caption{(a) Distribution of the rank $k$ where the relative diagonal magnitude $|r_{ii}|/|r_{11}|$ drops below $\zeta$, derived from SDXL drift embeddings across different thresholds. The dashed lines indicate the chosen bounds $[3, 15]$. (b) Mean CLIP Score achieved by ODC as a function of $\zeta$, with the red marker indicating the chosen threshold ($\zeta=1.0\%$).}
  \label{fig:rank_distribution}
\end{figure}

To justify the empirical selections in our pipeline, we conduct an ablation study on the ODC hyperparameters using the held-out 100-prompt subset of \textsc{Coal}Bench-300 described in Section~\ref{experiments}. The results are visualized in Figures~\ref{fig:hyperparameters} and~\ref{fig:rank_distribution}.

\textbf{Number of Keywords ($k$).} We evaluate the number of top retrieved vocabulary terms used to construct the orthogonal drift vector $\vect{v}_{\text{orth}}$. As shown in Figure~\ref{fig:hyperparameters}(a, d), performance saturates rapidly from top-1 to top-3. For FLUX.1, $k=3$ yields the clear peak performance. For SDXL, $k=3$ and $k=5$ perform nearly identically. However, we select $k=5$ for SDXL as it provides a better balance of semantic coverage. We therefore adopt $k=5$ for SDXL and $k=3$ for FLUX.1 as our final parameters.

\textbf{Concept Removal Weight ($\alpha$).} This parameter scales the magnitude of the correction. Figure~\ref{fig:hyperparameters}(b, e) demonstrates that optimal removal strength is model-dependent. SDXL requires a stronger correction ($\alpha=1.0$), whereas FLUX.1 is highly sensitive, as an $\alpha$ of 0.8 causes catastrophic over-correction and collapse (CLIP 24.957). We select $\alpha=0.3$ for FLUX.1, which provides a safe margin from the over-correction cliff while achieving strong alignment.

\textbf{Keywords Weights Temperature ($\tau$).} The temperature controls the sharpness of the softmax applied to the similarity scores. We select $\tau=0.50$ for SDXL and $\tau=0.05$ for FLUX.1. As shown in Figure~\ref{fig:hyperparameters}(c, f), ODC is robust to this parameter across an order of magnitude ($\tau \in [0.05, 1.0]$), confirming that the method does not rely on sharp, one-hot weighting.

\textbf{Adaptive Rank-Reduction Bounds and Energy Threshold ($k_{\min}$, $k_{\max}$, $\zeta$).} The adaptive rank-reduction in $\mathcal{R}$ (Algorithm~\ref{alg:adaptive_removal}) relies on bounds $k_{\min}=3$ and $k_{\max}=15$ to balance concept completeness against noise inclusion. To justify these bounds and our energy threshold $\zeta=0.01$, we analyze the drift embeddings $\mathbf{E}_{\text{orth}}$ across our validation set. As shown in Figure~\ref{fig:rank_distribution}(a), we computed the QR decomposition and recorded the rank $k$ where the relative diagonal magnitude $|r_{ii}|/|r_{11}|$ dropped below $\zeta$, prior to applying any clamping. Evaluated across multiple thresholds ($\zeta \in \{0.1\%, 0.5\%, 1.0\%, 2.0\%\}$), the rank consistently falls within the $[3, 15]$ range. Ranks below 3 are too sparse to represent a complete concept, while ranks above 15 encode irrelevant noise. 

Furthermore, Figure~\ref{fig:rank_distribution}(b) demonstrates that the threshold $\zeta=1.0\%$ provides the optimal empirical performance, yielding the highest CLIP Score (31.613). Tighter thresholds (e.g., $0.1\%$) or looser thresholds (e.g., $2.0\%$) incorporate either too much noise or too little semantic information, respectively. Therefore, $\zeta=1.0\%$ combined with the safety rails $[3, 15]$ ensures our concept removal operator targets only the unwanted semantic drift.

\section{Adaptation of Attend-and-Excite for SDXL}
\label{app:ae_details}

To employ Attend-and-Excite (A\&E) as a baseline for the Stable Diffusion XL (SDXL) architecture, we adapted the official implementation to account for the model's increased latent resolution and dual text-encoder structure. While the core Generative Semantic Nursing (GSN) logic remains consistent with \cite{chefer2023attend}, our implementation introduces the following specific changes:

\begin{enumerate}[leftmargin=*]
    \item \textbf{Attention Map Extraction:} The original method operates on $16 \times 16$ attention maps. In SDXL, we found this resolution (corresponding to a 64-pixel stride) too coarse. We instead extract maps from the $32 \times 32$ cross-attention layers of the U-Net. We aggregate attention probabilities across heads and derive token indices using the tokenizer associated with the second text encoder (OpenCLIP ViT-bigG/14).
    
    \item \textbf{Padding and Smoothing:} To prevent the optimization from exploiting edge artifacts, we apply \textit{reflection padding} to the attention maps prior to Gaussian smoothing (kernel size $3 \times 3$, $\sigma=0.5$). This ensures that subjects generated near the image boundaries are not penalized by zero-padding in the loss calculation. Additionally, we scale the raw attention scores by a factor of 100 before the Softmax operation to sharpen the attention distribution.

    \item \textbf{Optimization Schedule:} We perform the latent optimization during the first 25 steps of a 30-step generation schedule. We utilize an iterative refinement strategy at timesteps $t \in \{0, 10, 20\}$ with acceptance thresholds $\tau \in \{0.1, 0.5, 0.8\}$. The latent code $z_t$ is updated with a gradient step size $\alpha_t$ that follows a linear decay schedule:
    \begin{equation}
        \alpha_t = S \cdot \left(1 - 0.5 \cdot \frac{t}{T_{total}}\right),
    \end{equation}
    where the initial scale factor $S$ is set to 20.0. Gradients are normalized if their norm exceeds $1\times10^{-5}$ to maintain stability within the SDXL latent space.
    
    \item \textbf{Subject Selection:} We automate the selection of subject tokens using a Part-of-Speech (POS) tagger to identify nouns and proper nouns within the input prompts.
\end{enumerate}

\section{Statistical Significance Analysis}
\label{app:significance}

To assess the robustness of our reported improvements, we compute 95\% confidence intervals (CIs) and perform paired t-tests. To account for repeated measurements across seeds, we first average the scores across the 3 seeds for each prompt and then conduct our significance tests over the same \textsc{Coal}Bench-300 prompts used for our main results in Table~\ref{tab:main_results} (N=200). The complete results are detailed in Table~\ref{tab:significance}.

\begin{table}[htbp]
  \centering
  \caption{Statistical significance analysis across the 200-prompt \textsc{Coal}Bench-300 evaluation subset (N=200). We report Mean $\pm$ Std and 95\% CIs for the baseline, ODC, and ODC BoT, along with the paired t-test results for the ODC improvement.}
  \label{tab:significance}
  \resizebox{\textwidth}{!}{%
  \begin{tabular}{@{}lcccccc@{}}
    \toprule
    \textbf{Model} & \textbf{Metric} & \textbf{Baseline (Mean $\pm$ Std)} & \textbf{ODC (Mean $\pm$ Std)} & \textbf{ODC BoT (Mean $\pm$ Std)} & \textbf{$\Delta$ (95\% CI)} & \textbf{$p$-value} \\
    \midrule
    \multirow{2}{*}{SDXL} 
      & CLIP $\uparrow$ & 30.328 $\pm$ 3.226 & 31.798 $\pm$ 3.213 & 32.291 $\pm$ 3.043 & 1.470 {[}1.165, 1.774{]} & \textbf{$< 10^{-17}$} \\
      & BLIP $\uparrow$ & 0.6078 $\pm$ 0.0074 & 0.6121 $\pm$ 0.0067 & 0.6127 $\pm$ 0.0064 & 0.0042 {[}0.0036, 0.0048{]} & \textbf{$< 10^{-28}$} \\
    \addlinespace
    \multirow{2}{*}{FLUX.1} 
      & CLIP $\uparrow$ & 29.450 $\pm$ 3.386 & 29.918 $\pm$ 3.224 & 30.455 $\pm$ 3.227 & 0.468 {[}0.239, 0.698{]} & \textbf{$< 10^{-4}$} \\
      & BLIP $\uparrow$ & 0.6076 $\pm$ 0.0072 & 0.6084 $\pm$ 0.0071 & 0.6097 $\pm$ 0.0069 & 0.0008 {[}0.0004, 0.0012{]} & \textbf{$< 10^{-3}$} \\
    \bottomrule
  \end{tabular}%
  }
\end{table}

Our analysis confirms reliable improvements in alignment scores, with robust gains observed for both CLIP Score (SDXL: $p < 10^{-17}$, FLUX.1: $p < 10^{-4}$) and BLIP Score (SDXL: $p < 10^{-28}$, FLUX.1: $p < 10^{-3}$). Crucially, across all configurations, the 95\% confidence intervals for the performance delta $\Delta$ ($\text{ODC} - \text{Baseline}$) lie strictly above zero, demonstrating that ODC consistently improves alignment without compromising baseline fidelity. This statistical robustness is further corroborated by our human evaluation (Table~\ref{tab:human_eval}), where domain experts consistently favored ODC-corrected images.

\section{Quantitative Assessment of Visual Fidelity}
\label{app:kid}

To verify that ODC preserves visual fidelity, we conduct a quantitative evaluation using the Kernel Inception Distance (KID)~\citep{binkowski2018demystifying} and a human evaluation. First, we use KID to measure the distributional similarity between the images generated by ODC and the unmodified baseline. As an unbiased estimator of the squared Maximum Mean Discrepancy (MMD), KID can yield slightly negative values due to sampling variance; in practice, a score near zero indicates that the two distributions are statistically indistinguishable. As summarized in Table~\ref{tab:kid_fidelity}, the near-zero or marginally negative scores observed for both SDXL and FLUX.1 confirm that ODC does not degrade visual fidelity.

\begin{table}[htbp]
  \centering
  \small
  \begin{minipage}[t]{0.52\textwidth}
    \centering
    \caption{Visual fidelity assessment via KID Score (Mean $\pm$ Std) $\downarrow$. Near-zero or negative scores indicate no measurable degradation.}
    \label{tab:kid_fidelity}
    \renewcommand{\arraystretch}{1.2}
    \setlength{\tabcolsep}{4pt}
    \begin{tabular}{@{}lcc@{}}
      \toprule
      \textbf{Model} & CoalBench-300 & PartiPrompts \\
      \midrule
      SDXL & $0.00093 \pm 0.00023$ & $0.00034 \pm 0.00015$ \\
      FLUX.1 [schnell] & $-0.00152 \pm 0.00008$ & $-0.00095 \pm 0.00011$ \\
      \bottomrule
    \end{tabular}
  \end{minipage}%
  \hfill
  \begin{minipage}[t]{0.45\textwidth}
    \centering
    \caption{Human Evaluation of visual fidelity. Win rates are computed over decisive votes (ties excluded).}
    \label{tab:human_eval_fidelity}
    \renewcommand{\arraystretch}{1.2}
    \setlength{\tabcolsep}{4pt}
    \begin{tabular}{@{}lcc@{}}
      \toprule
      \textbf{Metric} & \textbf{Value} & \textbf{95\% CI} \\
      \midrule
      ODC Win Rate & 43.7\% & [35.3\%, 52.4\%] \\
      Baseline Win Rate & 56.3\% & [47.6\%, 64.7\%] \\
      \midrule
      Tie Rate & 58.0\% & --- \\
      \bottomrule
      \multicolumn{3}{l}{\footnotesize Binomial test $p=0.93$, Cohen's $h=0.13$ (negligible).} \\
    \end{tabular}
  \end{minipage}
\end{table}

Second, we conducted a human evaluation specifically targeting visual fidelity. Three domain experts performed 300 blind pairwise comparisons between baseline and ODC-corrected images. The results show a high tie rate of 58.0\%, indicating that ODC does not introduce perceptible artifacts. Furthermore, the win rates in decisive votes (Method: 43.7\%, Baseline: 56.3\%) show no statistically significant difference (Binomial test $p=0.93$, Cohen's $h=0.13$). This confirms that ODC preserves visual fidelity without measurable degradation, corroborating the KID findings. These results validate the core design of ODC. The method functions as a surgical modification to the conditioning space, successfully removing semantic drift components while ensuring the generated images remain firmly within the manifold of natural image distributions.

\clearpage

\section{Additional Qualitative Results}
\label{app:qualitative_appendix}

In this section, we provide further qualitative results to complement the main paper. These examples demonstrate our method's ability to improve prompt-image alignment and preserve visual fidelity across a diverse set of prompts. 

\begin{figure}[htbp]
    \centering
    \includegraphics[width=.66\textwidth]{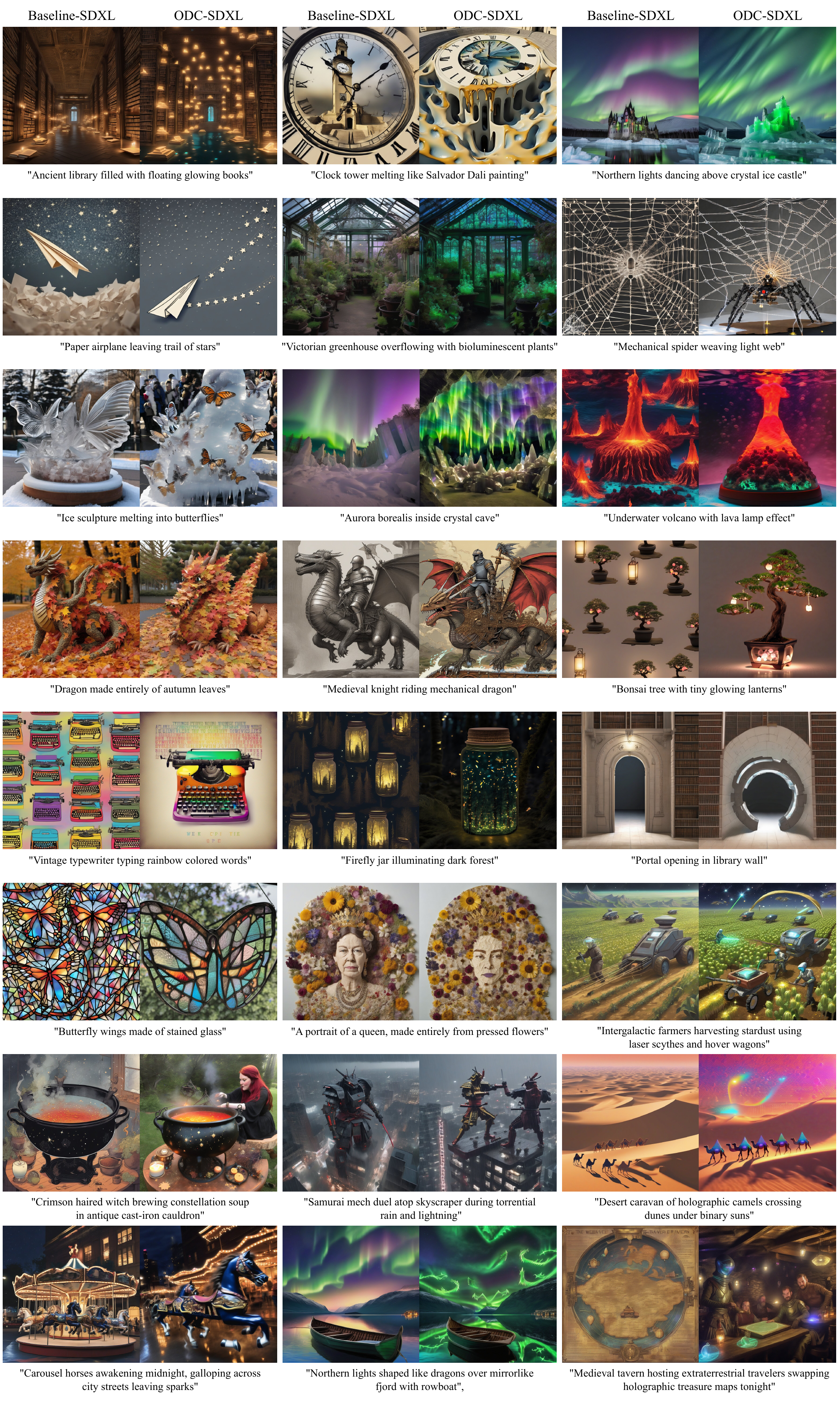}
    \caption{Additional qualitative results comparing baseline generations with Orthogonal Drift Correction (ODC). For each prompt, we show the output from the baseline SDXL model (left) and our ODC-corrected version (right).}
    \label{fig:qualitative_sdxl_appendix}
\end{figure}

\clearpage

\begin{figure}[htbp]
    \centering
    \includegraphics[width=.66\textwidth]{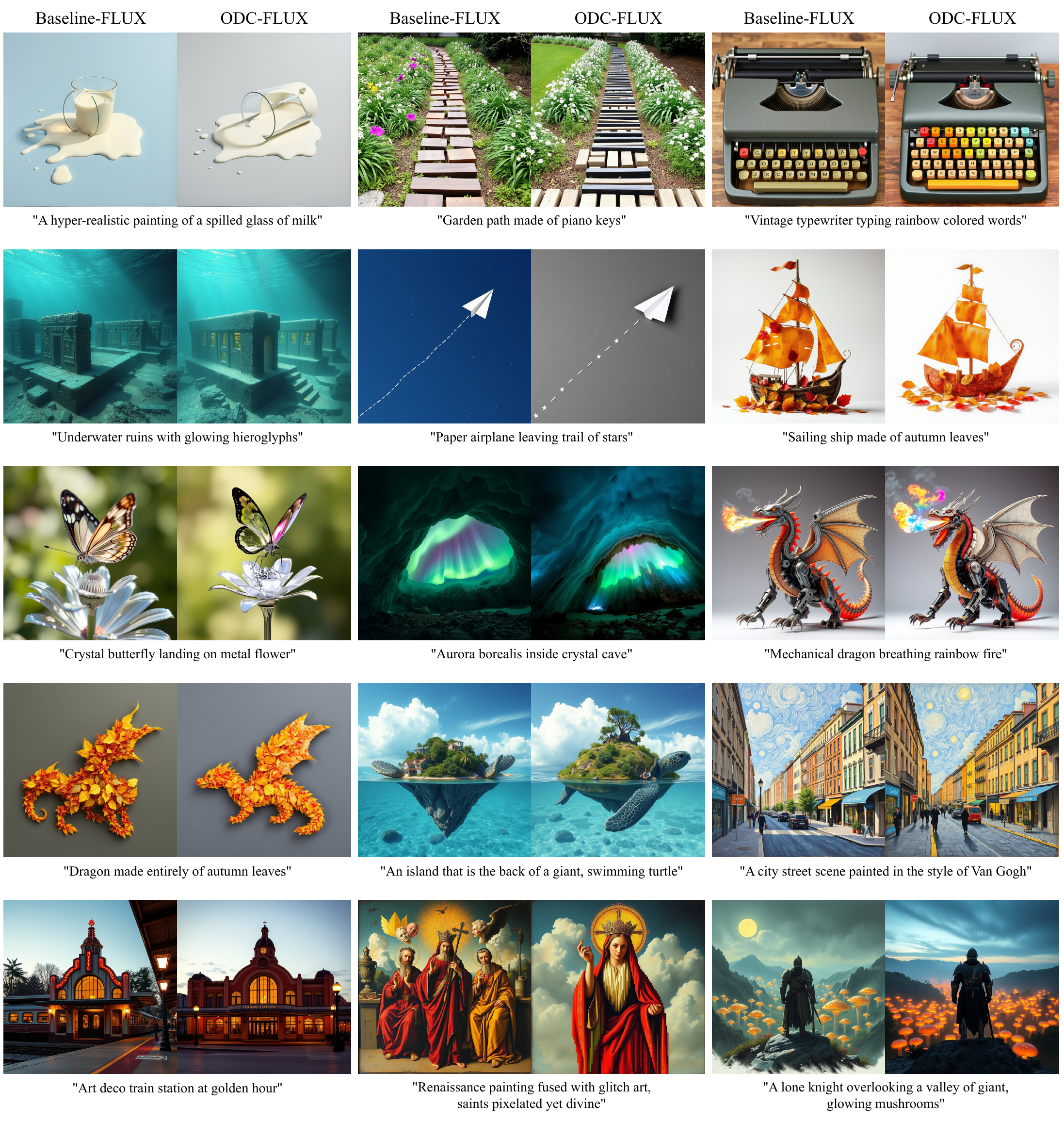}
    \caption{Additional qualitative results comparing baseline generations with Orthogonal Drift Correction (ODC). For each prompt, we show the output from the baseline FLUX.1 model (left) and our ODC-corrected version (right).}
    \label{fig:qualitative_flux_appendix}
\end{figure}

\section{Retrieved Drift Vocabulary $\mathcal{S}_{\text{orth}}$}
\label{app:drift_words}

To provide examples of the retrieved vocabulary used by ODC, we list the top-5 terms ($\mathcal{S}_{\text{orth}}$) for the exact prompts featured in our qualitative results (Figure~\ref{fig:qualitative_comparison}) in Table~\ref{tab:drift_words}. These terms serve as the discrete textual surrogate for the isolated orthogonal drift vector ($\vect{v}_{\text{orth}}$). ODC utilizes the embeddings of these top-5 terms in aggregate to define the removal direction. Consequently, while individual words may appear abstract in isolation, their combined embedding closely approximates the true semantic drift direction, allowing ODC to target the spurious concepts accurately.

\begin{table}[htbp]
  \centering
  \footnotesize 
  \caption{Retrieved Drift Vocabulary ($\mathcal{S}_{\text{orth}}$) for Representative Prompts}
  \label{tab:drift_words}
  \renewcommand{\arraystretch}{1.3} 
  \begin{tabularx}{\textwidth}{@{} c >{\raggedright\arraybackslash}X c >{\raggedright\arraybackslash}X >{\raggedright\arraybackslash}X @{}}
    \toprule
    \textbf{Index} & \textbf{Prompt Text} & \textbf{Model} & \textbf{Retrieved Drift Words ($\mathcal{S}_{\text{orth}}$)} & \textbf{Interpretation} \\
    \midrule
    151 & A tree whose leaves are made of delicate, colored glass & SDXL & \texttt{stained glass art}, \texttt{glass art}, \texttt{light panels}, \texttt{light panel}, \texttt{glasswork} & Unintended glass/art style elements. \\
    149 & A medieval feast with ghosts sitting amongst the living guests & SDXL & \texttt{the abbess}, \texttt{illustre}, \texttt{preillustration}, \texttt{overillustration}, \texttt{tabernacler} & Misplaced religious/illustrative terms. \\
    30 & Paint brushes growing like flowers in pot & SDXL & \texttt{utensils or brushes}, \texttt{craft materials}, \texttt{brush land}, \texttt{petrographer}, \texttt{makeup kit} & Spurious, unrelated tools/materials. \\
    4 & Cherry blossoms falling on samurai armor & SDXL & \texttt{wallpaper}, \texttt{rider and horse}, \texttt{person horse}, \texttt{illustre}, \texttt{art wall decor} & Misplaced compositional elements. \\
    46 & Giant mushroom houses in enchanted forest & SDXL & \texttt{mushroom by}, \texttt{forest floor}, \texttt{agaricic}, \texttt{agaricales}, \texttt{toader} & Unexpected botanical elements. \\
    63 & Clockwork owl perched on moon & SDXL & \texttt{tartary}, \texttt{overillustration}, \texttt{odontolith}, \texttt{nightfowl}, \texttt{antecavern} & Unrelated archaic/anatomical terms. \\
    280 & Candy-coated castle melting under sunrise... & SDXL & \texttt{enchanted landscape}, \texttt{castle on}, \texttt{fantasy scene}, \texttt{rapunzel}, \texttt{castle} & Generic fantasy tropes overriding details. \\
    203 & Victorian astronaut playing violin inside ornate greenhouse... & SDXL & \texttt{the duo}, \texttt{musicians}, \texttt{a duet}, \texttt{twopenny}, \texttt{the two brothers} & Erroneous secondary subject. \\
    300 & Thunderstorm brewing inside snow globe... & SDXL & \texttt{in the winter}, \texttt{house in the}, \texttt{with snow}, \texttt{semiglobe}, \texttt{snow globe} & Literal winter concepts overriding interior. \\
    \midrule
    10 & Northern lights dancing above crystal ice castle & FLUX.1 & \texttt{iglulirmiut}, \texttt{alpenglow}, \texttt{the arctic ocean}, \texttt{arcticwards}, \texttt{icelidae} & Unexpected arctic/celestial concepts. \\
    24 & Giant teacup island in cloudy sky & FLUX.1 & \texttt{islandlike}, \texttt{that island}, \texttt{land formation}, \texttt{an island}, \texttt{a small island} & Detail misinterpreted as geography. \\
    185 & A collection of magical potions, glowing on a dusty shelf & FLUX.1 & \texttt{liquor bottles}, \texttt{small bottles}, \texttt{alcohol bottles}, \texttt{bottles of liquor}, \texttt{bottle liquor} & Mundane domain overriding fantasy context. \\
    85 & Floating geometric islands at sunset & FLUX.1 & \texttt{these islands}, \texttt{some islands}, \texttt{many islands}, \texttt{air lane sunrise}, \texttt{small islands} & Repetitive island concepts. \\
    31 & Lighthouse made of stacked books & FLUX.1 & \texttt{lighthouse picture}, \texttt{lighthouse}, \texttt{light house}, \texttt{lighthouse base}, \texttt{the lighthouse} & Redundant literal subject terms. \\
    158 & A mountain range shaped like a sleeping stone dragon & FLUX.1 & \texttt{the western alps}, \texttt{the alps}, \texttt{the eastern alps}, \texttt{the italian alps}, \texttt{the fra mauro formation} & Geographic literalism over metaphor. \\
    291 & Lavender fields beneath glass dome... & FLUX.1 & \texttt{beekeeping}, \texttt{lavender plants}, \texttt{apiary}, \texttt{pollenlike}, \texttt{the pollen} & Real-world agriculture over sci-fi. \\
    150 & An underwater city inhabited by elegant, humanoid sea creatures & FLUX.1 & \texttt{the new atlantis}, \texttt{undersea}, \texttt{the ocean floor}, \texttt{underseas}, \texttt{the sea bottom} & Environmental setting overriding architectural subject. \\
    268 & Digital renaissance fair where knights duel using AR dragons & FLUX.1 & \texttt{dragonlike}, \texttt{a dragon}, \texttt{their armor}, \texttt{dragon statue}, \texttt{their armour} & Literal dragons/armor over AR concept. \\
    \bottomrule
  \end{tabularx}
\end{table}

\end{document}